\documentclass{article}

\usepackage[preprint,nonatbib]{neurips_2020}

\usepackage[utf8]{inputenc} %
\usepackage[T1]{fontenc}    %
\usepackage{hyperref}       %
\usepackage{url}            %
\usepackage{booktabs}       %
\usepackage{amsfonts}       %
\usepackage{nicefrac}       %
\usepackage{microtype}      %
\usepackage{graphicx}
\usepackage{tabularx}
\usepackage{enumitem}

\usepackage{tikz}
\usepackage{comment}
\usepackage{amsmath,amssymb} %
\usepackage{color}
\usepackage{float}
\DeclareMathOperator*{\argmin}{arg\,min}
\DeclareMathOperator*{\argmax}{arg\,max}
\DeclareMathOperator{\Tr}{Tr}
\newcommand{\pgrad}[1]{\ensuremath{\frac{\partial L}{\partial {#1}}}}

\title{An Analysis of SVD for Deep Rotation Estimation}

\author{
\begin{tabular*}{0.95\textwidth}{c@{\extracolsep{\fill}}c@{\extracolsep{\fill}}c@{\extracolsep{\fill}}c}
	\hspace{0.8cm}Jake Levinson\textsuperscript{1} &
	Carlos Esteves\textsuperscript{2} &
	Kefan Chen\textsuperscript{3} &
	Noah Snavely\textsuperscript{3} \\
	\multicolumn{4}{c}{\begin{tabular}{c@{\hspace*{.85cm}}c@{\hspace*{.85cm}}c} Angjoo Kanazawa\textsuperscript{3} &
	Afshin Rostamizadeh\textsuperscript{3} &
	Ameesh Makadia\textsuperscript{3}\end{tabular}}
\end{tabular*}\\
\begin{tabular*}{0.85\textwidth}{c@{\extracolsep{\fill}}c@{\extracolsep{\fill}}c}
	\textsuperscript{1}University of Washington &
	\textsuperscript{2}University of Pennsylvania &
	\textsuperscript{3}Google Research
\end{tabular*}\\
}

\begin{document}

\maketitle

\begin{abstract}
Symmetric orthogonalization via SVD, and closely related procedures, are well-known techniques for projecting matrices onto $O(n)$ or $SO(n)$. These tools have long been used for applications in computer vision, for example optimal 3D alignment problems solved by orthogonal Procrustes, rotation averaging, or Essential matrix decomposition. Despite its utility in different settings, SVD orthogonalization as a procedure for producing rotation matrices is typically overlooked in deep learning models, where the preferences tend toward classic representations like unit quaternions, Euler angles, and axis-angle, or more recently-introduced methods. 
Despite the importance of 3D rotations in computer vision and robotics, a single universally effective representation is still missing. Here, we explore the viability of SVD orthogonalization for 3D rotations in neural networks. We present a theoretical analysis that shows SVD is the natural choice for projecting onto the rotation group. Our extensive quantitative analysis shows simply replacing existing representations with the SVD orthogonalization procedure obtains state of the art performance in many deep learning applications covering both supervised and unsupervised training.
\end{abstract}

\section{Introduction}
There are many ways to represent a 3D rotation matrix. But what is the ideal representation to predict 3D rotations in a deep learning framework? The goal of this paper is to explore this seemingly low-level but practically impactful question, as currently the answer appears to be ambiguous.

In this paper we present a systematic study on estimating rotations in neural networks. We identify that the classic technique of SVD orthogonalization, widely used in other contexts but rarely in the estimation of 3D rotations in deep networks, is ideally suited for this task with strong empirical and theoretical support. 

3D rotations are important quantities appearing in countless applications across different fields of study, and are now especially ubiquitous in learning problems in 3D computer vision and robotics. The task of predicting 3D rotations is common to estimating object pose~\cite{xiang2017posecnn,mahendran17cvprw,MousavianCVPR17,su15iccv,vpsKpsTulsianiM15,li18eccv,sundermeyer2018eccv}, relative camera pose~\cite{melekhov17acivs,poursaeed2018eccv,en18eccvw}, ego-motion and depth from video~\cite{zhou2017unsupervised,mahjourian18cvpr}, and human pose~\cite{zhou2016deep,hmrKanazawa17}.

A design choice common to all of these models is selecting a representation for 3D rotations. The most frequent choices are classic representations including unit quaternion, Euler angles, and axis-angle. Despite being such a well-studied problem, there is no universally effective representation or regression architecture due to performance variations across different applications.%

A natural alternative to these classic representations is symmetric orthogonalization, a long-known technique which projects matrices onto the orthogonal group $O(3)$~\cite{lowdin70aqs,procrustes}. Simple variations can restrict the projections onto the special orthogonal (rotation) group $SO(3)$~\cite{Horn:88,kabsch,wahba}. This procedure, when executed by Singular Value Decomposition (SVD~\cite{golubbook}), has found many applications in computer vision, for example at the core of the Procrustes problem~\cite{arunprocrustes,procrustes} for point set alignment, as well as single rotation averaging~\cite{hartley13ijcv}. A nearly identical procedure is used for factorizing Essential matrices~\cite{mvg03book}.

Despite its adoption in these related contexts, orthogonalization via SVD has not taken hold as a procedure for generating 3D rotations in deep learning: it is rarely used when implementing a model (e.g.\ overlooked in~\cite{li18eccv,en18eccvw,poursaeed2018eccv,mahjourian18cvpr}), nor is it considered a benchmark when evaluating new representations~\cite{s2reg,hao6d}.

In light of this, this paper explores the viability of SVD orthogonalization for estimating rotations in deep neural networks. Note, we do not claim to be the first to introduce this tool to deep learning, rather our focus is on providing a comprehensive study of the technique specifically for estimating rotations. Our contributions include
\begin{itemize}[leftmargin=1cm,topsep=0pt]%
    \item A theoretically motivated analysis of rotation estimation via SVD orthogonalization in the context of neural networks, and in comparison to the recently proposed Gram-Schmidt procedure~\cite{hao6d}. One main result is that SVD improves over Gram-Schmidt by a factor of two for reconstruction, thus supporting SVD as the preferred orthogonalization procedure.
    \item An extensive quantitative evaluation of SVD orthogonalization spanning four diverse application environments: point cloud alignment, object pose from images, inverse kinematics, and depth prediction from images, across supervised and unsupervised settings, and benchmarked against classic and recently introduced rotation representations.
\end{itemize}

Our results show that rotation estimation via SVD orthogonalization achieves state of the art performance in almost all application settings, and is the best performing method among those that can be applied in both supervised and unsupervised settings. This is an important result given the prevalence of deep learning frameworks that utilize rotations, as well as for benchmarking future research into new representations.

\section{Related Work}
Optimization on $SO(3)$, and more generally on Riemannian manifolds, is a well-studied problem. Peculiarities arise since $SO(3)$ is not topologically homeomorphic to any subset of 4D Euclidean space, so any parameterization in four or fewer dimensions will be discontinuous (this applies to all classic representations---Euler angles, axis-angle, and unit quaternions). 
Discontinuities and singularities are a particular nuisance for classic gradient-based optimization on the manifold~\cite{stuelpnagel64rotations,cjtaylorSO3}.

Early deep learning models treated Euler angle estimation as a classification task~\cite{vpsKpsTulsianiM15,su15iccv}, by discretizing the angles into bins and using softmax to predict the angles. This idea was extended to hybrid approaches that combine classification and regression. In~\cite{rcnn3d}, discrete distributions over angles are mapped to continuous angles via expectation and \cite{classificationregression2018} combines classification over quantized rotations with the regression of a continuous offset. In~\cite{s2reg} it is shown that typical activations used in classification models (e.g.\ softmax) lead to more stable training compared to the unconstrained setting of regression. The authors introduce a ``spherical exponential'' mapping to bridge the gap and improve training stability for regression to $n$-spheres. All methods that discretize suffer from the problem of increased dimensionality and expressivity.  All of the above methods 
require supervision on the classification objective, which makes 
them unsuitable for 
unsupervised settings.

Probabilistic representations have been introduced for modeling orientation with uncertainty~\cite{deepdirectstat2018,Gilitschenski2020Deep}, with the von Mises and Bingham distributions respectively. While these are best suited for multi-modal and ambiguous data, such approaches do not reach state of the art in tasks where a single precise rotation must be predicted.

The closest approach to SVD orthogonalization is the recent work of~\cite{hao6d} which makes a strong connection between the discontinuities of a representation and their effect in neural networks. In search of continuous representations, they propose the idea of continuous overparameterizations of $SO(3)$, followed by continuous projections onto $SO(3)$. Their $6D$ representation is mapped onto $SO(3)$ via a partial Gram-Schmidt procedure. This is similar in spirit to SVD orthogonalization which will map a continuous 9D representation onto $SO(3)$ with SVD. We leave the deeper comparison of these two methods to the following sections, where we show that SVD provides a more natural projection onto $SO(3)$ in several respects (Section~\ref{sec:analysis}). %

SVD derivatives have been presented in~\cite{giles2008cmd,papad00svd}, and there exist multiple works that build neural nets with structured layers depending on components of SVD or Eigendecomposition~\cite{ionesculong,probst2019cvpr,Huang_2018_CVPR,dang18eccv,miyato2018spectral}. Close to our setting are~\cite{suwajanakorn2018discovery} which applies the orthogonal Procrustes problem to 3D point set alignment within a neural network, and~\cite{jia19pami} which proposes singular value clipping to regularize networks' weight matrices. 

SVD is amenable for learning via backpropagation. Its derivatives have been presented in~\cite{giles2008cmd,papad00svd}, and there exist multiple works that build neural nets with structured layers depending on SVD or Eigendecomposition~\cite{ionesculong,probst2019cvpr,Huang_2018_CVPR,dang18eccv}. The closest to our setting is~\cite{suwajanakorn2018discovery} which applies the orthogonal Procrustes problem to 3D point set alignment within a neural network, and~\cite{jia19pami} which proposes singular value clipping to regularize networks' weight matrices. We discuss the stability of SVD orthogonalization in neural networks in the following section.

\section{Analysis}\label{sec:analysis}
In this section we present our theoretically motivated analysis of SVD orthogonalization for rotation estimation. We start here defining the procedures and introducing well-known results regarding their least-squares optimality before presenting the analysis.

Given a square matrix $M$ with SVD $U\Sigma V^T$, we consider the orthogonalization and \emph{special} orthogonalization
\begin{align}
\texttt{SVDO}(M) &:= UV^T, \\
\texttt{SVDO}^+(M) &:= U \Sigma' V^T, \text{ where } \Sigma' = \mathrm{diag}(1, \ldots, 1, \det(UV^T)). \label{svdplus}
\end{align}
$\texttt{SVDO}$ is orientation-preserving, while $\texttt{SVDO}^+$ maps to $SO(n)$. Orthogonalization of a matrix via SVD is also known as \emph{symmetric orthogonalization}~\cite{lowdin70aqs}.
It is well known that symmetric orthogonalization is optimal in the least-squares sense ~\cite{arunprocrustes,Horn:88,procrustes}:
\begin{align}
    \texttt{SVDO}(M) = \argmin_{R \in O(n)} \lVert R - M\rVert_F^2, %
    \qquad
    \texttt{SVDO}^+(M) = \argmin_{R \in SO(n)} \lVert R - M\rVert_F^2. \label{least-squares}%
\end{align}
This property has made symmetric orthogonalizations useful in a variety of applications~\cite{wahba,procrustes,suwajanakorn2018discovery}.

To reiterate, the procedure for deep rotation estimation we will evaluate experimentally in Section~\ref{sec:experiments} is $\texttt{SVDO}^+(M)$ in 3D, which takes a 9-dimensional network output (interpreted as a $3 \times 3$ matrix $M$), and projects it onto $SO(3)$ following Eq.~\ref{svdplus}.

\subsection[Maximum likelihood estimates]{$\texttt{SVDO}(M)$ and $\texttt{SVDO}^+(M)$ are maximum likelihood estimates}
\label{subsec:MLE}
In this section we show that SVD orthogonalization maximizes the likelihood and minimizes the expected error in the presence of Gaussian noise.
Let $M = R_\mu + \sigma N, M\in \mathbb{R}^{n \times n}$ represent an observation of $R_\mu \in SO(n)$, corrupted by noise $N$ with entries $n_{ij} \sim \mathcal{N}(0, 1)$. 
Following from the pdf of the matrix normal distribution~\cite{gupta1999matrix} the likelihood function is
\begin{equation}\small
    L(R_\mu; M, \sigma) =
    ((2\pi)^{\frac{n^2}{2}}\sigma^{n^2})^{-1}
    \exp({-\tfrac{1}{2\sigma^2}((M - R_\mu)^T (M - R_\mu))})
    \label{mle}.
\end{equation}
$L(R_\mu; M, \sigma)$, subject to $R_\mu \in SO(n)$, is maximized when $(M - R_\mu)^T (M - R_\mu)$ is minimized:
\begin{equation}\small
    \argmax_{R_\mu \in SO(n)} L(R_\mu; M, \sigma) = \argmin_{R_\mu \in SO(n)} (M - R_\mu)^T (M - R_\mu) = \argmin_{R_\mu \in SO(n)} \lVert M - R_\mu\rVert_F^2
\end{equation}
The solution is given by $\texttt{SVDO}^+(M)$ (Eq.~\ref{least-squares}). Similarly, $\texttt{SVDO}(M)$ will minimize Eq.~\ref{mle} when $R_\mu \in O(n)$.

\subsection{Gradients}\label{backprop}
In this section we analyze the behaviour of the gradients of a network with an $\texttt{SVDO}^+$ layer, and show that they are generally well behaved.  Specifically, we can consider \pgrad{M} for some a loss function $L(M, R) = \lVert \texttt{SVDO}^+(M) - R \rVert_F^2$. We will first analyze \pgrad{M} for $\texttt{SVDO}(M)$.
Letting $\circ$ denote the Hadamard product, from~\cite{ionesculong,townsendsvd}
we have%
\begin{eqnarray}\small
\pgrad{M} & = & U [(F \circ (U^T \pgrad{U} - \pgrad{U}^T U))\Sigma + \Sigma ( F \circ (V^T \pgrad{V} - \pgrad{V}^T V))]V^T, \label{gradstart}\\
F_{i,j} & = & 
\begin{cases}
    \frac{1}{s_i^2 - s_j^2},& \text{if } i \neq j \label{invsig}\\
    0,& \text{if } i = j
\end{cases}, \;\;s_i = \Sigma_{ii}
\end{eqnarray}
Letting $X = U^T \pgrad{U} - \pgrad{U}^T U$, and $Y = V^T \pgrad{V} - \pgrad{V}^T V$,  we see that $X, Y$ are antisymmetric and $X = -Y$ (this is a result of the loss function depending on $UV^T$). We can reduce $\pgrad{M} = UZV^T$ where the elements of $Z$ are
\begin{eqnarray}\small
Z_{ij} & = &
\begin{cases}
    \frac{-X_{ij}}{s_i + s_j} ,& \text{if } i \neq j \label{invsig2}\\
    0,& \text{if } i = j.
\end{cases}
\end{eqnarray}
See Appendix~\ref{sec:appgrads} for the detailed derivation.
For $\texttt{SVDO}(M)$ Eq.~\ref{invsig2} tells us $\pgrad{M}$ is undefined whenever two singular values are both zero and large when their sum is very near zero. 

For $\texttt{SVDO}^+(M)$, if $\det(M) > 0$ then the analysis is the same as above. If $\det(M) < 0$, the extra factor $D = \mathrm{diag}(1, 1, \ldots, -1)$ effectively changes the smallest singular value $s_n$ to $-s_n$. The derivation is otherwise unchanged. In particular the denominator in equation \eqref{invsig2} is now $s_j - s_n$ or $s_n - s_i$ if either $i$ or $j$ is $n$. Thus, $\pgrad{M}$ is undefined if the smallest singular value occurs with multiplicity greater than $1$. It is large if the two smallest singular values are close to each other, or if they are close to $0$.

\subsection{Error analysis}\label{svd_vs_gs}
In this section we approximate the expected error in $\texttt{SVDO}(M)$ and Gram-Schmidt orthogonalization (denoted as $\texttt{GS}(M)$) in the presence of Gaussian noise, and observe that the error is twice as large for $\texttt{GS}$ as for $\texttt{SVDO}$. 
If $M$ is a matrix with QR decomposition $M = QR$, define:
\begin{equation}
\texttt{GS}(M) := Q, \qquad \texttt{GS}^+(M) := Q \Sigma'', \text{ where } \Sigma'' = \mathrm{diag}(1, \ldots, 1, \det(Q)).\label{gsplus}
\end{equation}
We consider $M = R_0+\sigma N$, a noisy observation of a rotation matrix $R_0 \in SO(n)$, where $N$ has i.i.d. Gaussian entries $n_{ij} \sim \mathcal{N}(0, 1)$ and $\sigma$ is small. Since $\det(M) > 0$ as $\sigma \to 0$, the analysis is identical for $\texttt{SVDO}^+$ and $\texttt{GS}^+$. The analysis is also independent of $R_0$ (Appendix~\ref{sec:propcorrfullproof}), so for simplicity we set $R_0 = I$. First we calculate the SVD and QR decompositions of $M$ to first order for $N$ an arbitrary (non-random) matrix.

\newtheorem{prop}{Proposition}
\newtheorem{cor}{Corollary}
\begin{prop}
The SVD and QR decompositions of $M = I + \sigma N$ are as follows:
\begin{itemize}
\item[1.] (SVD) Let $N = S+A$ be the decomposition of $N$ into symmetric and antisymmetric parts. Then, to first order, an SVD of $M$ is given by
\[M = U_0(I+\sigma U_1) \cdot (I + \sigma \Sigma_1) \cdot (I + \sigma V_1)^T U_0^T,\]
where $U_0 \Sigma_1 U_0^T$ is an SVD of $S$, and $U_1, V_1$ are (non-uniquely determined) antisymmetric matrices satisfying $U_0^T A U_0 = U_1 + V_1^T$.
\item[2.] (QR) Let $N = U + D + L$ be the strict upper-triangular, diagonal, and strict lower-triangular parts of $N$. To first order, $M$ has QR decomposition
\begin{align*}
M &= (I + \sigma Q_1) \cdot (I + \sigma R_1), %
\end{align*}
where $Q_1 = L - L^T$ and $R_1 = D + U + L^T$.
\end{itemize}
Consequently, $\emph{\texttt{SVDO}}(M) = I + \sigma A + O(\sigma^2)$ and $\emph{\texttt{GS}}(M) = I + \sigma(L - L^T) + O(\sigma^2)$.
\end{prop}

\begin{cor}
If $N$ is $3 \times 3$ with i.i.d. Gaussian entries $n_{ij} \sim \mathcal{N}(0, 1)$, then with error of order $O(\sigma^3)$,

\begin{alignat}{2}
\mathbb{E}[\lVert \emph{\texttt{SVDO}}(M) - I \rVert_F^2] &= 3\sigma^2,\qquad
\mathbb{E}[\lVert \emph{\texttt{GS}}(M) - I \rVert_F^2] &&= 6\sigma^2 \\
\mathbb{E}[\lVert \emph{\texttt{SVDO}}(M) - M \rVert_F^2] &= 6\sigma^2,\qquad
\mathbb{E}[\lVert \emph{\texttt{GS}}(M) - M\rVert_F^2] &&= 9\sigma^2
\end{alignat}
\end{cor}
This corollary is by straightforward calculation since each entry of $A$ and $L - L^T$ is Gaussian (see appendix~\ref{sec:propcorrfullproof} for the proof). Notably, Gram-Schmidt produces \emph{twice} the error in expectation (and indeed deviates $1.5$ times further from the observation $M$ itself). The same holds for $\texttt{SVDO}^+$ and $\texttt{GS}^+$. This difference in performance can be traced to the fact that Gram-Schmidt is essentially "greedy" with respect to the starting matrix, whereas the SVD approach is coordinate-independent.

Although i.i.d.\ Gaussian noise is not necessarily reflective of a neural network's predictions, it does provide meaningful insight into the relationship between $\texttt{SVDO}^+$ and $\texttt{GS}^+$ and their relative performance observed in practice.  See Appendix~\ref{sec:propcorrfullproof} for further remarks.  

\emph{Proof of Proposition (sketch; see Appendix~\ref{sec:propcorrfullproof} for the full proof).}

(1) Let $M$ have SVD $M = U S V^T$ for some orthogonal matrices $U, V$ and diagonal matrix $S \geq 0$. To first order in $\sigma$, we can write
\begin{equation}
    I + \sigma N = U_0(I + \sigma U_1)(S_0 + \sigma S_1)(I + \sigma V_1)^T V_0^T
\end{equation}
with $U_0, V_0$ orthogonal, $U_1, V_1$ antisymmetric and $S_0, S_1 \geq 0$ diagonal. (This is using the fact that the antisymmetric matrices give the tangent space to the orthogonal matrices.) The claims come from breaking this equation into symmetric and antisymmetric parts; note if $X$ is (anti-)symmetric and $Q$ is orthogonal, then $QXQ^T$ is again (anti-)symmetric. Simplify to get $\texttt{SVDO}(M) = I + \sigma A + O(\sigma^2).$

For (2) the proof is similar: we expand the QR decomposition $M = QR$ to first order as
\begin{equation}
I+\sigma N = Q_0(I+\sigma Q_1)(I+\sigma R_1)R_0,
\end{equation}
$Q_0$ orthogonal, $Q_1$ antisymmetric, and $R_1, R_0$ upper triangular; examine the upper and lower parts.

\subsection{Continuity for special orthogonalization}\label{continuity}
The calculation above shows $\texttt{SVDO}(M)$ and $\texttt{SVDO}^+(M)$ are continuous and differentiable, at least at $M=I$. In fact $\texttt{SVDO}(M)$ is smooth, as is $\texttt{SVDO}^+$ except for a discontinuity\footnote{If $f$ is "discontinuous on a set $S$" of measure 0, it is equivalently "continuous on $\mathbb{R}^n \setminus S$."} if (and only if) $\det(M) = 0$ or $\det(M) < 0$ and its smallest singular value has multiplicity greater than 1. In fact the optimization problem \eqref{least-squares} is degenerate in this case. For example, the $2 \times 2$ matrix $M = \mathrm{diag}(1, -1)$ is equidistant from every rotation matrix; perturbations of $M$ may special-orthogonalize to any $R \in SO(2)$. %
$\texttt{GS}^+$ is continuous on a slightly larger domain -- $\det(M) \ne 0$ -- because it makes a uniform choice, negating the $n$-th column of $M$ if necessary, at the cost of significantly greater error in expectation. This reflects the fact that SVD orthogonalization is coordinate-independent and $\texttt{GS}, \texttt{GS}^+$ are not:
\begin{equation}
    \texttt{SVDO}(R_1 M R_2) = R_1 \texttt{SVDO}(M)R_2, \text{ for all } R_1, R_2 \in SO(n), M \in GL(n),
\end{equation}
and similarly for $\texttt{SVDO}^+$. $\texttt{GS}$ and $\texttt{GS}^+$ are rotation-equivariant on only one side: $\texttt{GS}(R_1M) = R_1 \texttt{GS}(M)$, but $\texttt{GS}(MR_2)$ is not a function of $R_2$ and $\texttt{GS}(M)$; likewise for $\texttt{GS}^+$. See Appendix~\ref{sec:appsmoothness} for a proof of smoothness and further discussion.

\subsection{Summary}
The results above illustrate a number of desirable properties of SVD orthogonalization. $\texttt{SVDO}^+$ is optimal as a projection in the least squares sense as well as in the presence of Gaussian noise (MLE). Viewed through the lens of matrix reconstruction, the approximation errors are half that of the Gram-Schmidt procedure. Finally we show the conditions that lead to large gradient norms (conditions that are rare for small matrices).  In the following, we support this theoretical analysis with extensive quantitative evaluations.

\section{Experiments}\label{sec:experiments}
This section presents the experimental analysis for SVD orthogonalization and numerous baseline methods. To recap, the SVD orthogonalization procedure $\texttt{SVDO}^+(M)$ takes a 9D network output (interpreted as a $3 \times 3$ matrix), and projects it onto $SO(3)$ via Eq.~\ref{svdplus}. The procedure can easily be used in popular deep learning libraries (e.g. PyTorch~\cite{pytorch} and TensorFlow~\cite{tensorflow} both provide differentiable SVD ops), and it does not affect efficiency ($3 \times 3$ SVD adds negligible computational overhead in both forward and backward passes).

\paragraph{Methods.} Now we provide a short description of the methods under comparison (see Appendix~\ref{sec:appbaselines} for further details).
\textbf{SVD-Train} is $\texttt{SVDO}^+(M)$ (Eq.~\ref{svdplus}).
\textbf{SVD-Inference} is $\texttt{SVDO}^+(M)$, except the training loss is applied directly to $M$. Since $\texttt{SVDO}^+$ is applied only at inference, it is a continuous representation for training.
\textbf{6D} and \textbf{5D} are introduced in~\cite{hao6d} for projecting 6D and 5D representations onto $SO(3)$. 6D is equivalent to $\texttt{GS}^+(M)$ (Eq.~\ref{gsplus}), and 5D utilizes a stereographic projection.
\textbf{Spherical Regression}~\cite{s2reg} (\textbf{$S^2$-Reg}) combines regression to the absolute values of a unit quaternion with classification of the signs. 
\textbf{3D-RCNN}~\cite{rcnn3d} combines likelihood estimation and regression for predicting Euler angles.
\textbf{Geodesic-Bin-and-Delta} (\textbf{M$_G$}~\cite{classificationregression2018}) presents a hybrid model which combines classification over quantized pose space with regression of offsets from the quantized poses.
\textbf{Quaternion}, \textbf{Euler angles}, and \textbf{axis-angle} are the classic parameterizations. In each case they are converted to matrix form before the loss is applied as in~\cite{hao6d}.

For SVD, 6D, 5D, and the classic representations, the loss is $L(R, R_{t}) = \frac{1}{2} \left \lVert R-R_t \right \rVert_{F}^2$. When $R, R_t \in SO(3)$ this is related to geodesic angle error $\theta$ as $L(R, R_t) = 2 - 2 \cos(\theta)$. All other methods require an additional classification loss. See Appendix~\ref{sec:appexp} for additional experimental details (architectures, implementations, etc).

\subsection{3D point cloud alignment}\label{sec:expptcloud1}
The first experiment is the point cloud alignment benchmark from~\cite{hao6d}. Given two shape point clouds the network is asked to predict the 3D rotation that best aligns them. The rotations in the dataset are sampled uniformly from $SO(3)$ (no rotation bias in the data).%
Table~\ref{tab:PTstats} (left) shows geodesic error statistics (mean, median, std) on the test set. We omit the maximum error as it is approx. $180^\circ$ for all methods, a reflection of the symmetries in the data rather than limitations of the methods. SVD-Train outperforms all the baselines, and even SVD-Inference performs on par with the best baseline (6D). Interestingly, the hybrid approaches 3D-RCNN and $M_G$ underperform the top regression baselines, a point we will return to later.
Table~\ref{tab:PTstats} (middle) shows the mean errors on the test set as training progresses. The best performing methods at the end of training (SVD variations, and 6D) also show fast convergence.  The errors at different percentiles are shown in Table~\ref{tab:PTstats} (right).

\begin{table}
\caption{\small\textbf{3D point cloud alignment.} Left: a comparison of methods by \emph{mean}, \emph{median}, and \emph{standard deviation} of (geodesic) errors after 2.6M training steps. Middle: mean test error at different points along the training progression. Right: test error percentiles after training completes. The legend on the right applies to both plots.}\label{tab:PTstats}
\centering
\resizebox{0.3\textwidth}{!}{
\begin{tabular}{l@{\hskip 0.05in}c@{\hskip 0.04in}c@{\hskip 0.07in}c}
& Mean ($^\circ$) & Med & Std \\\cline{2-4}
3D-RCNN & 5.51 & 1.91 & 17.05 \\
$M_G$ & 9.12 & 7.65 & 10.46 \\
Euler & 11.04 & 6.23 & 15.56 \\
Axis-Angle & 6.65 & 4.06 & 11.47 \\
Quaternion & 5.48 & 3.19 & 11.03 \\
$S^2$-Reg & 4.80 & 3.00 & 9.27 \\
5D & 3.77 & 2.19 & 8.70 \\
6D & 2.24 & 1.22 & 7.83 \\
SVD-Inf & 2.64 & 1.60 & 8.16 \\
SVD-Train & \textbf{1.63} & \textbf{0.89} & \textbf{6.70}
\end{tabular}
}
\begin{tabular}{ll}
\hspace{-1em}
\raisebox{-0.2in}{\includegraphics[width=0.34\textwidth]{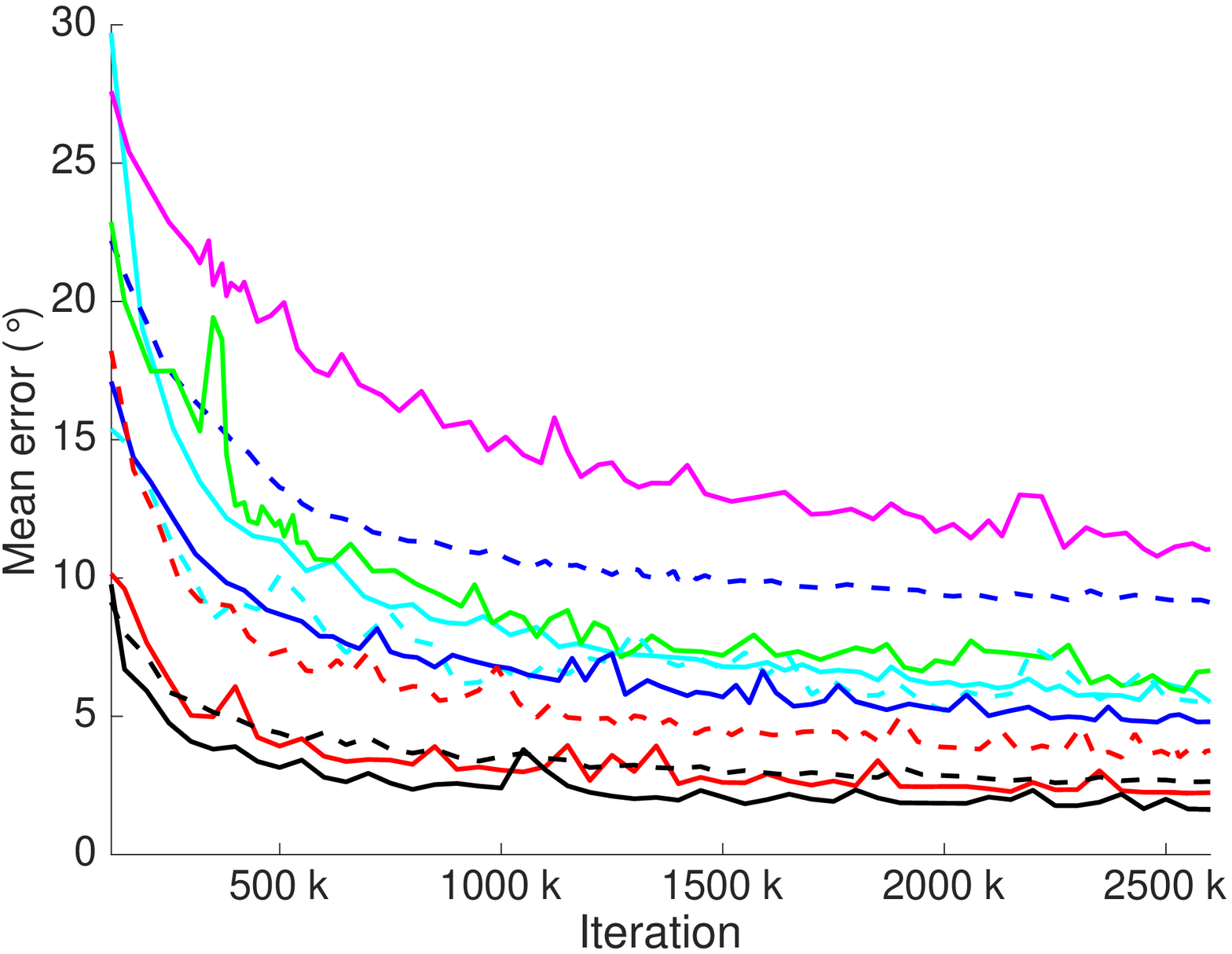}} &
\hspace{-1.5em}
\raisebox{-0.2in}{\includegraphics[width=0.34\textwidth]{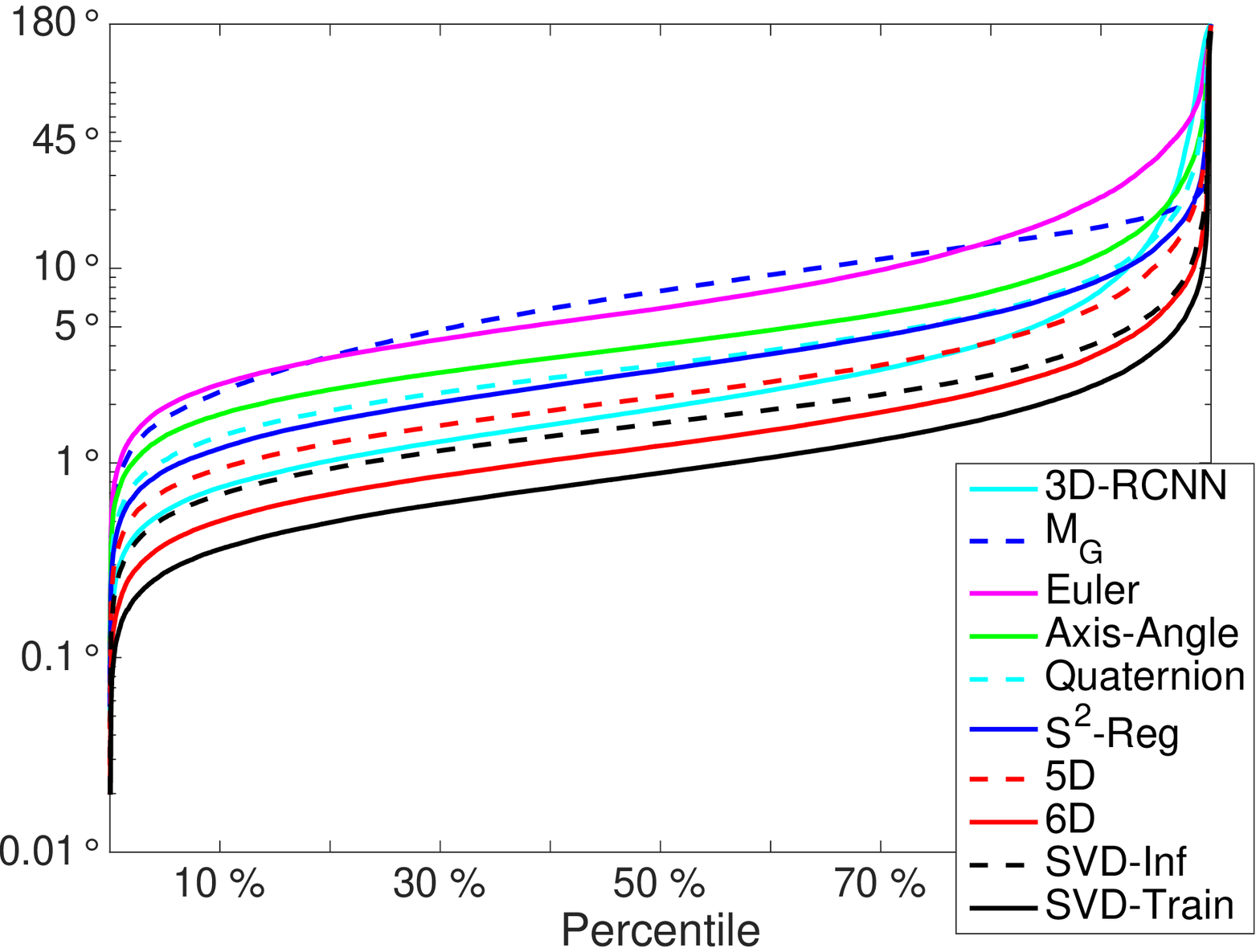}}
\end{tabular}
\end{table}

\subsection{3D Pose estimation from 2D images}\label{sec:mnet}
The second experiment follows the benchmark set forth in~\cite{s2reg}. Images are rendered from ModelNet10~\cite{modelnet} objects from arbitrary viewpoints. Given a 2D image, the network must predict the object orientation. We used MobileNet~\cite{howard2017mobilenets} to generate image features, followed by fully connected regression layers. Rather than averaging over all 10 ModelNet categories as in~\cite{s2reg}, we focus on \emph{chair} and \emph{sofa} which are the two categories which exhibit the least rotational symmetries in the dataset. 
Results are shown in Tables~\ref{tab:MNetChairStats} and~\ref{tab:MNetSofaStats}. Interestingly, SVD-Inference also performs similarly to SVD-Train on final metrics with faster convergence, indicating short pretraining with SVD-Inference could improve convergence rates.

3D-RCNN and $M_G$ are again underperforming the best methods. These hybrid methods have shown state of the art performance on predicting 3D pose from images~\cite{rcnn3d,classificationregression2018}, but in those benchmarks the 3D rotations exhibit strong bias (camera viewpoints are not evenly distributed over $SO(3)$) which is a property their classification networks can exploit. In our experiments so far we have only considered rotations uniformly sampled from $SO(3)$.
\begin{table}
\caption{\small\textbf{Pose estimation from ModelNet chair images.} We report the same metrics as in Table~\ref{tab:PTstats}, see the caption there for a description. All models are trained for 550K steps in this case.}
\label{tab:MNetChairStats}
\centering
\resizebox{0.3\textwidth}{!}{
\begin{tabular}{l@{\hskip 0.05in}c@{\hskip 0.04in}c@{\hskip 0.07in}c}
& Mean ($^\circ$) & Med & Std \\\cline{2-4}
3D-RCNN & 35.50 & 13.21 & 46.55 \\
$M_G$ & 31.60 & 16.70 & 41.86 \\
Euler & 41.35 & 27.44 & 37.73 \\
Axis-Angle & 32.30 & 19.74 & 34.70 \\
Quaternion & 26.92 & 14.39 & 32.92 \\
$S^2$-Reg & 27.36 & 15.41 & 33.17 \\
5D & 25.18 & 13.40 & 32.10 \\
6D & 22.60 & 11.51 & 31.24 \\
SVD-Inf & 21.38 & 11.41 & \textbf{29.35} \\
SVD-Train & \textbf{21.25} & \textbf{11.14} & 30.28
\end{tabular}
}
\begin{tabular}{ll}
\hspace{-1em}
\raisebox{-0.2in}{\includegraphics[width=0.34\textwidth]{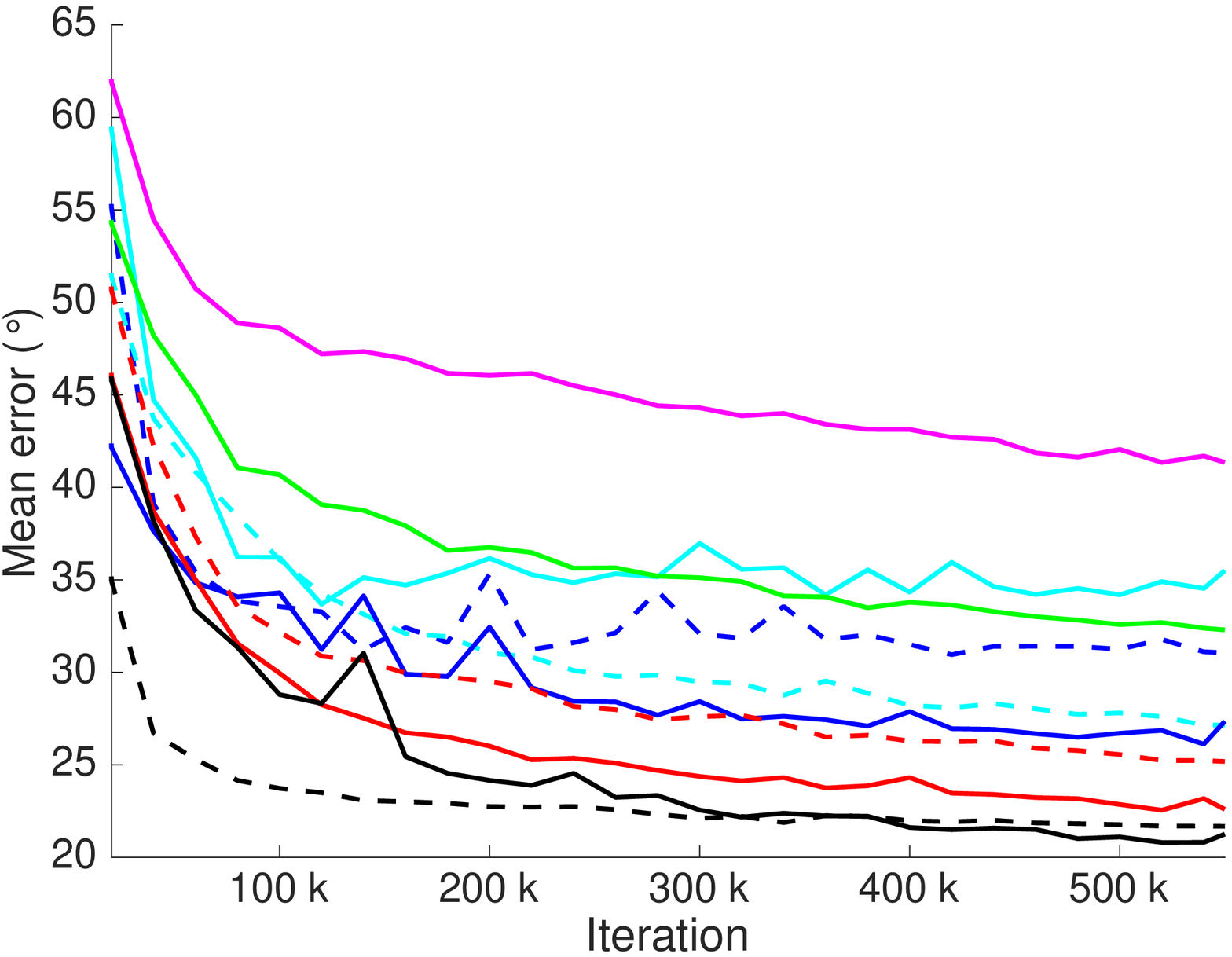}} &
\hspace{-1.5em}
\raisebox{-0.2in}{\includegraphics[width=0.34\textwidth]{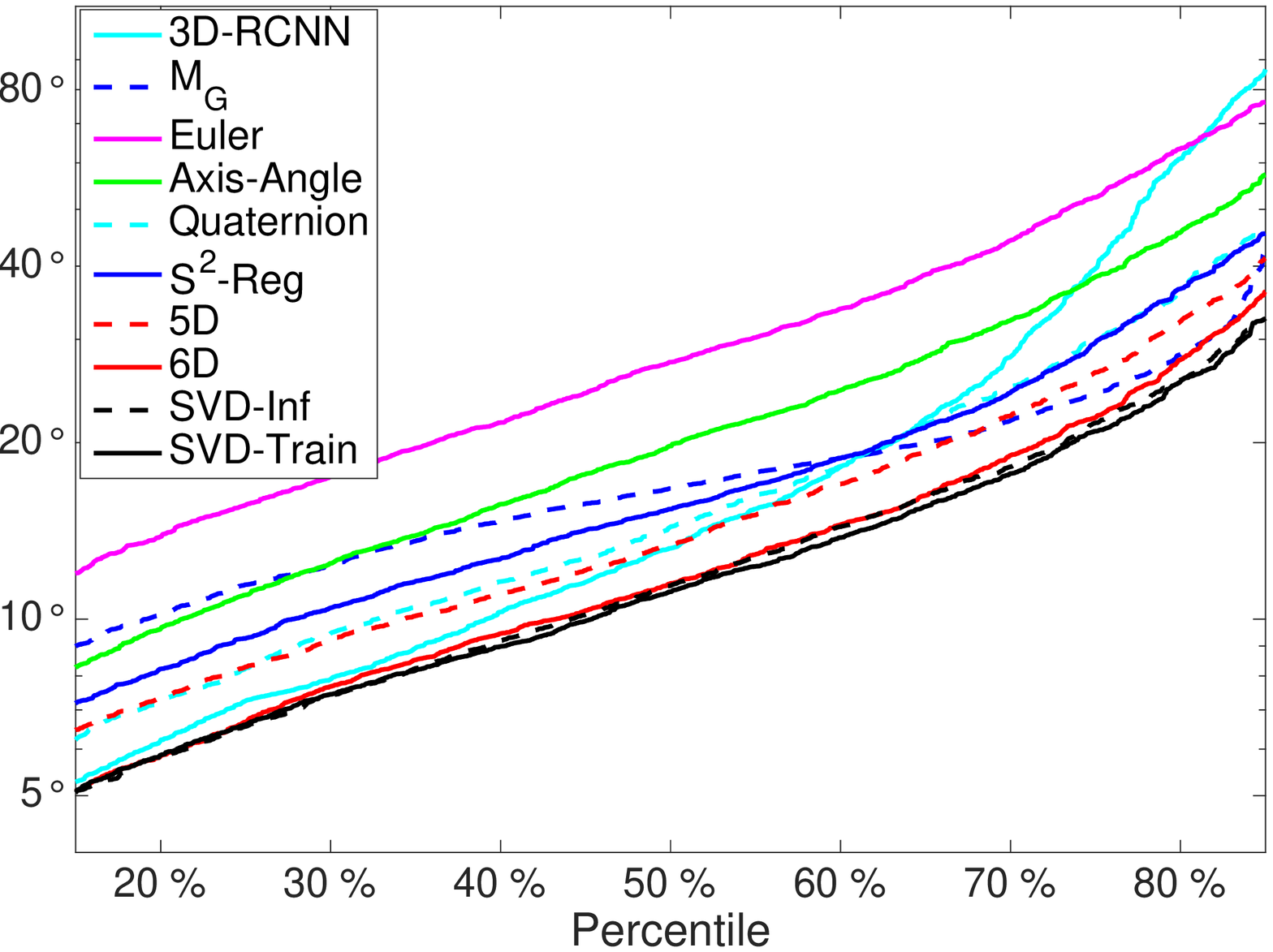}}
\end{tabular}
\end{table}

\begin{table}
\caption{\small\textbf{Pose estimation from ModelNet sofa images.} We report the same metrics as in Table~\ref{tab:PTstats}, see the caption there for a description. All models are trained for 550K steps in this case.}
\label{tab:MNetSofaStats}
\centering
\resizebox{0.3\textwidth}{!}{
\begin{tabular}{l@{\hskip 0.05in}c@{\hskip 0.04in}c@{\hskip 0.07in}c}
& Mean ($^\circ$) & Med & Std \\\cline{2-4}
3D-RCNN & 34.80 & 7.32 & 55.73 \\
$M_G$ & 31.41 & 13.93 & 48.48 \\
Euler & 49.31 & 32.03 & 43.47 \\
Axis-Angle & 31.82 & 17.31 & 37.19 \\
Quaternion & 29.60 & 14.56 & 37.00 \\
$S^2$-Reg & 25.99 & 12.11 & 37.67 \\
5D & 26.23 & 11.52 & 38.91 \\
6D & 20.25 & 7.84 & 36.85 \\
SVD-Inf & 20.30 & 8.85 & \textbf{33.88} \\
SVD-Train & \textbf{18.01} & \textbf{7.31} & 33.96
\end{tabular}
}
\begin{tabular}{ll}
\hspace{-1em}
\raisebox{-0.2in}{\includegraphics[width=0.34\textwidth]{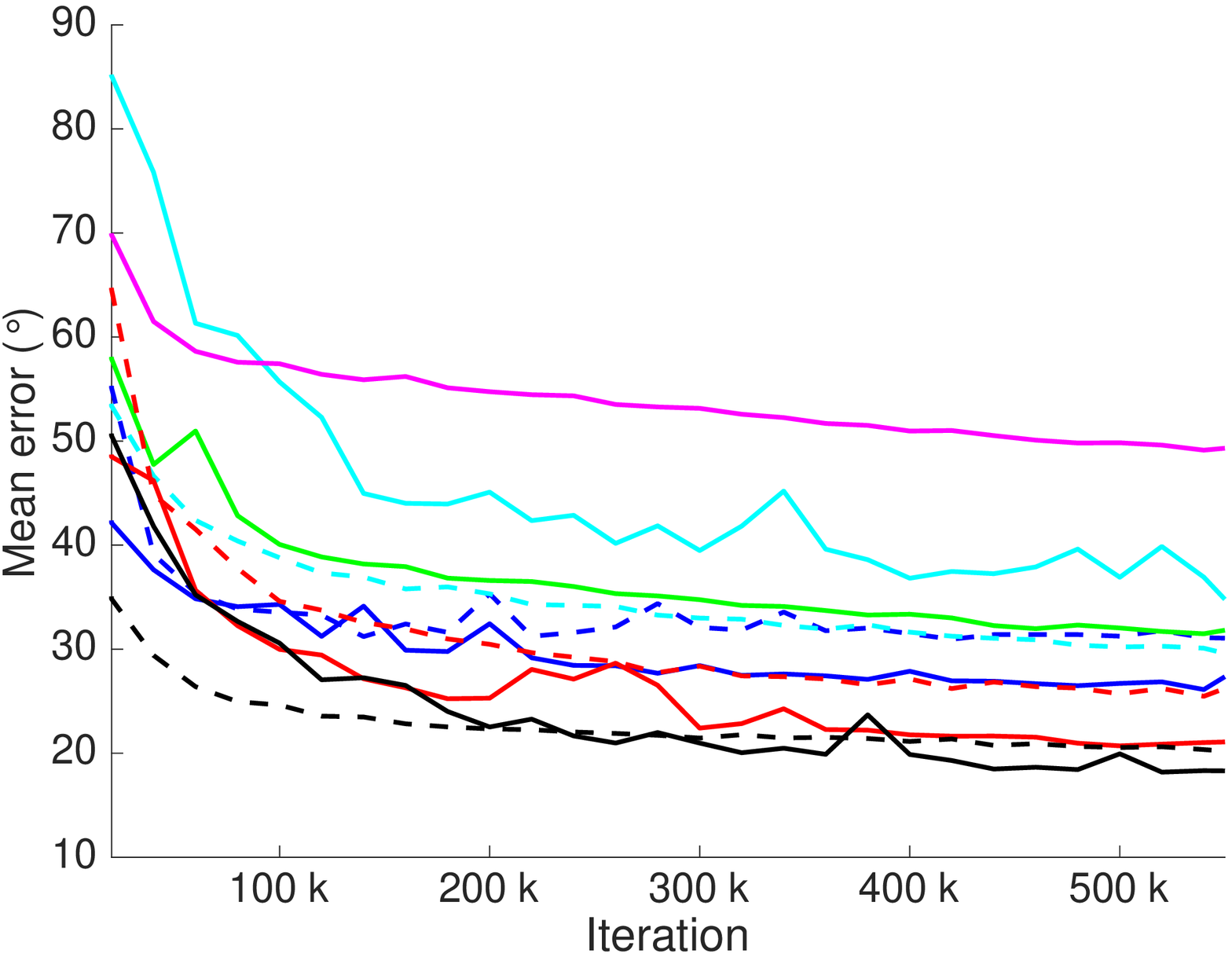}} &
\hspace{-1.5em}
\raisebox{-0.2in}{\includegraphics[width=0.34\textwidth]{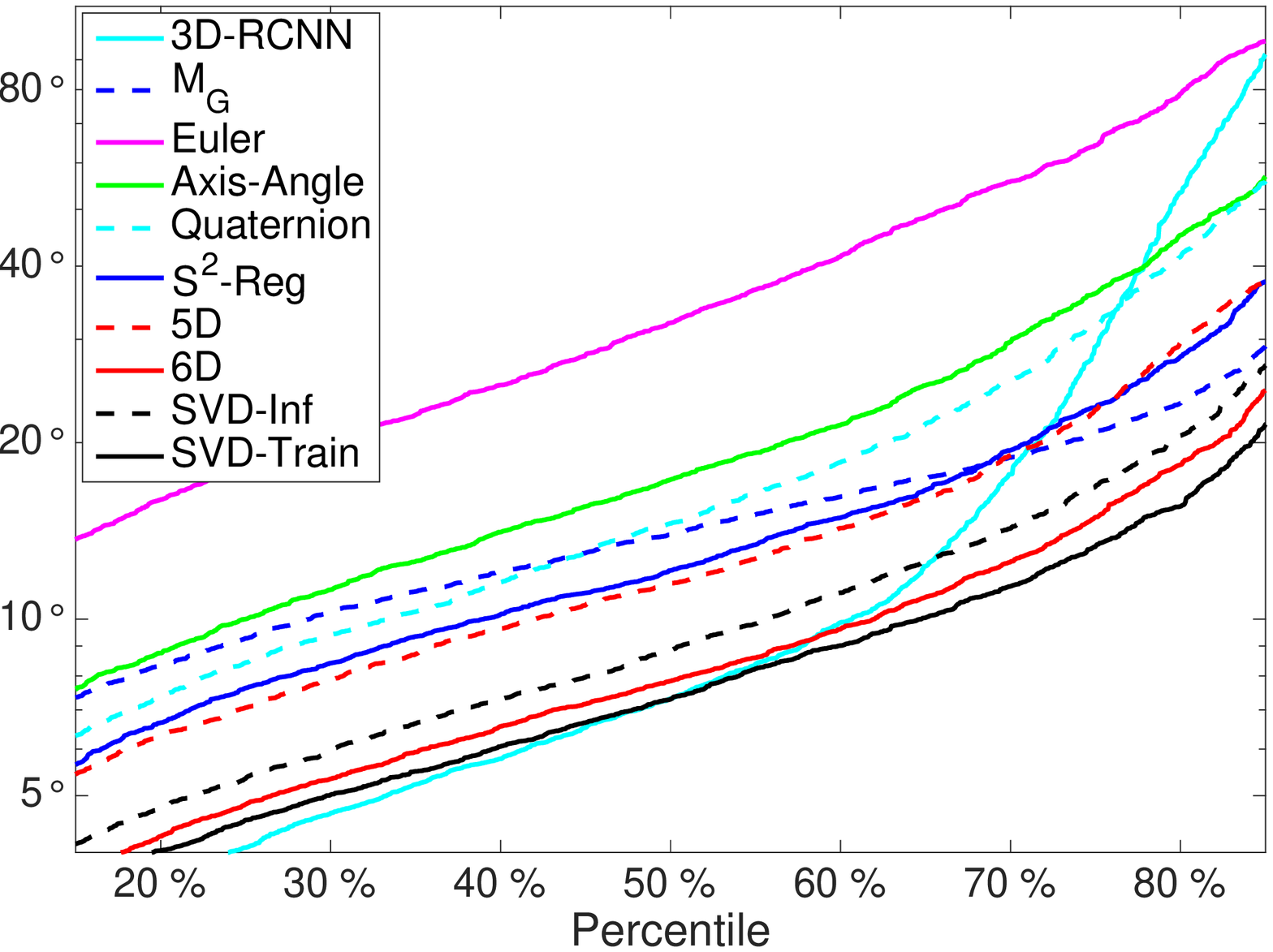}}
\end{tabular}
\end{table}

\subsection{Pascal 3D+}\label{sec:p3d}
Pascal3D+~\cite{pascal3d} is a standard benchmark for object pose estimation from single images. The dataset is composed of real images covering 12 categories. For training we discard occluded or truncated objects~\cite{classificationregression2018,su15iccv} and augment with rendered images from~\cite{su15iccv}. The model architecture is the same as Section~\ref{sec:mnet}.  Table~\ref{tab:pascal3d} shows results on two categories and the mean over all categories (see Appendix~\ref{sec:apppascal} for results on each of the 12 categories). The individual metrics we report are the median error as well as accuracies at $10^\circ$, $15^\circ$, and $20^\circ$.

The best performing method is clearly $S^2$-Reg. As expected, the hybrid method 3D-RCNN performs well on this task, but SVD-Inference and SVD-Train are on par. The SVD variations are also the best performing of the regression methods (those that only train with a rotation loss). Interestingly, SVD-Inference slightly outperforms SVD-Train, which suggests in this scenario where viewpoints have a non-uniform prior, training a network to regress directly to the desired rotation can work well.
\begin{table}
\caption{\textbf{Pascal 3D+.} Accuracy at $10^\circ$, $15^\circ$, and $20^\circ$ (higher is better), and median error are reported. On the left are results for \emph{sofa} and \emph{bicycle}. The third block is the results averaged over all 12 categories, and these numbers are used to determine the ranks shown on the right (lower is better).}
\label{tab:pascal3d}
\resizebox{\textwidth}{!}{
\centering
\small
\begin{tabular}{lcccccccccccccccc}
 & \multicolumn{4}{c}{Sofa} & \multicolumn{4}{c}{Bicycle} & \multicolumn{4}{c}{Mean (12 categories)} & \multicolumn{4}{c}{Rank (12 categories)}\\
 \cmidrule(lr){2-5}\cmidrule(lr){6-9}\cmidrule(lr){10-13}\cmidrule(lr){14-17}
 & \multicolumn{3}{c}{Accuracy@} &  Med$^\circ$ & \multicolumn{3}{c}{Accuracy@} &  Med$^\circ$ & \multicolumn{3}{c}{Accuracy@} & Med$^\circ$ & \multicolumn{3}{c}{Accuracy@} &  Med$^\circ$\\
 & 10$^\circ$ & 15$^\circ$ & 20$^\circ$ & Err & 10$^\circ$ & 15$^\circ$ & 20$^\circ$ & Err & 10$^\circ$ & 15$^\circ$ & 20$^\circ$ & Err & 10$^\circ$ & 15$^\circ$ & 20$^\circ$ & Err \\
\cmidrule(lr){2-4}\cmidrule(lr){5-5}\cmidrule(lr){6-8}\cmidrule(lr){9-9}\cmidrule(lr){10-12}\cmidrule(lr){13-13}\cmidrule(lr){14-16}\cmidrule(lr){17-17}
3D-RCNN & 37.1 & 54.3 & 80.0 & 14.2 & 17.8 & 38.6 & 72.3 & 16.9 & 43.2 & 57.6 & 78.1 & 12.9 & 2 & 3 & 5 & 2 \\
$M_G$ & 31.4 & 51.4 & 74.3 & 14.4 & 11.9 & 31.7 & 66.3 & 20.9 & 32.9 & 52.4 & 77.0 & 14.7 & 6 & 5 & 7 & 5 \\
Euler & 22.9 & 45.7 & 77.1 & 16.3 & 9.9 & 20.8 & 68.3 & 23.4 & 24.5 & 42.0 & 71.9 & 19.2 & 9 & 10 & 10 & 10 \\
Axis-Angle & 11.4 & 40.0 & 80.0 & 16.3 & 13.9 & 31.7 & 70.3 & 21.3 & 23.0 & 44.3 & 76.9 & 17.7 & 10 & 8 & 8 & 9 \\
Quaternion & 34.3 & 62.9 & 77.1 & 11.7 & 15.8 & 30.7 & 67.3 & 22.4 & 34.2 & 51.6 & 78.0 & 15.1 & 5 & 6 & 6 & 6 \\
$S^2$-Reg & 37.1 & \textbf{65.7} & 85.7 & 11.2 & \textbf{21.8} & \textbf{45.5} & 75.2 & \textbf{16.1} & \textbf{45.8} & \textbf{64.4} & \textbf{83.8} & \textbf{11.3} & \textbf{1} & \textbf{1} & \textbf{1} & \textbf{1} \\
5D & 17.1 & 54.3 & 77.1 & 14.2 & 10.9 & 26.7 & 68.3 & 21.1 & 25.2 & 43.9 & 75.6 & 17.0 & 8 & 9 & 9 & 8 \\
6D & 34.3 & 54.3 & \textbf{88.6} & 13.3 & 10.9 & 26.7 & 68.3 & 21.1 & 32.6 & 51.1 & 81.1 & 15.2 & 7 & 7 & 3 & 7 \\
SVD-Inf & \textbf{45.7} & 60.0 & \textbf{88.6} & \textbf{11.0} & 10.9 & 33.7 & \textbf{84.2} & 19.0 & 39.9 & 58.7 & 83.7 & 13.0 & 3 & 2 & 2 & 3 \\
SVD-Train & 40.0 & 57.1 & 85.7 & 12.7 & 9.9 & 26.7 & 80.2 & 20.9 & 35.1 & 52.7 & 80.5 & 14.6 & 4 & 4 & 4 & 4
\end{tabular}
}
\end{table}

\subsection{Unsupervised rotations}
So far we have considered supervised rotation estimation. Given the growing attention to self- or unsupervised 3D learning~\cite{zhou2017unsupervised,mahjourian18cvpr,hmrKanazawa17,hao6d}, it is important to understand how different representations fare without direct rotation supervision. We omit 3D-RCNN, $M_G$, and $S^2$-Reg from the experiments below as they require explicit supervision of classification terms, as well as SVD-Inference as it does not produce outputs on $SO(3)$ while training.

\subsubsection{Self-supervised 3D point cloud alignment}\label{sec:ptcldunsup}
To begin, we devise a simple variation of the point cloud alignment experiment from Section~\ref{sec:expptcloud1}. Given two point clouds, the network still predicts the relative rotation. However, now the only loss is L2 on the point cloud registration after applying the predicted rotation. All other experiment details remain the same. From Table~\ref{tab:PTIntStats}, SVD-Train performs significantly better than the next closest baseline, 6D.
\begin{table}
\caption{\small\textbf{Self-supervised 3D point cloud alignment.} The error metrics presented follow the same format as the earlier supervised point cloud alignment experiment, see Table~\ref{tab:PTstats}. Although here the model is trained without rotation supervision, we show test errors in the predicted rotations. The legend on the right applies to both plots.}
\label{tab:PTIntStats}
\centering
\resizebox{0.3\textwidth}{!}{
\begin{tabular}{l@{\hskip 0.05in}c@{\hskip 0.04in}c@{\hskip 0.07in}c}
& Mean ($^\circ$) & Med & Std \\\cline{2-4}
Euler & 12.83 & 6.68 & 18.72 \\
Axis-Angle & 6.90 & 4.06 & 12.12 \\
Quaternion & 5.76 & 3.19 & 12.19 \\
5D & 3.85 & 2.18 & 9.07 \\
6D & 2.39 & 1.35 & 7.78 \\
SVD-Train & \textbf{1.58} & \textbf{0.88} & \textbf{6.51}
\end{tabular}
}
\begin{tabular}{ll}
\hspace{-1em}
\raisebox{-0.2in}{\includegraphics[width=0.34\textwidth,height=1.3in]{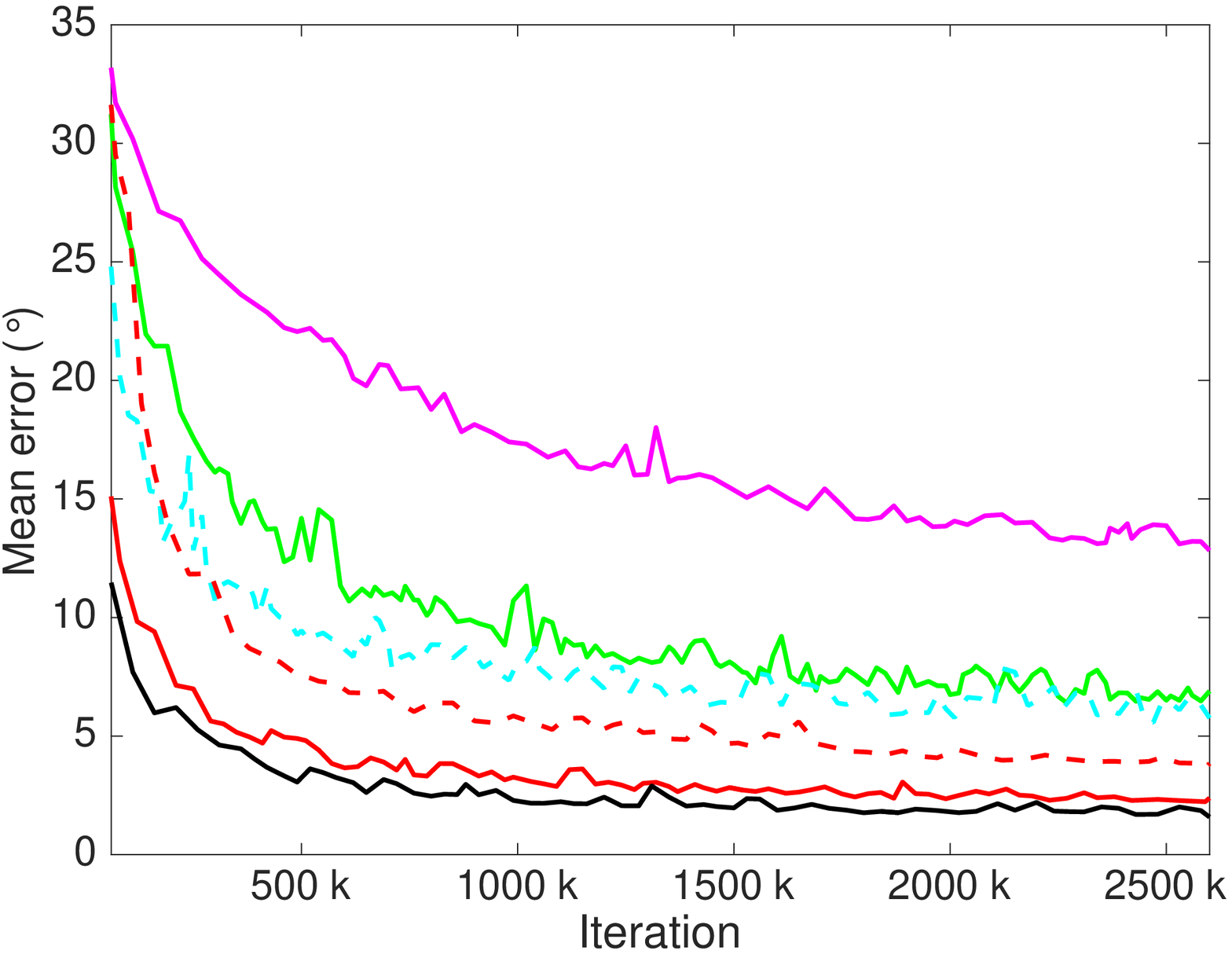}} &
\hspace{-1.5em}
\raisebox{-0.2in}{\includegraphics[width=0.34\textwidth,height=1.3in]{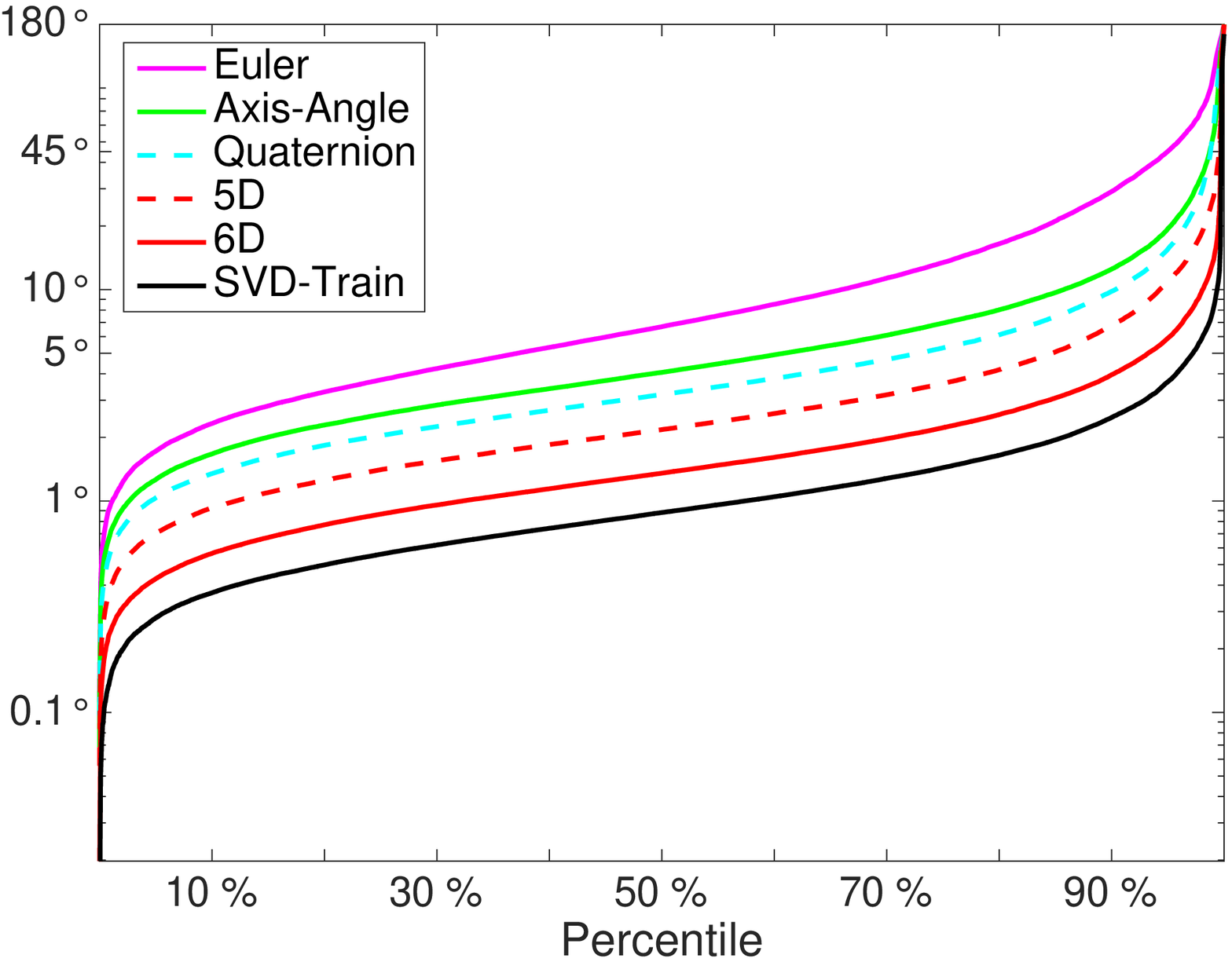}}
\end{tabular}
\end{table}
\subsubsection{Inverse kinematics}\label{sec:ik}
Our second unsupervised experiment is the human pose inverse kinematics experiment presented in~\cite{hao6d}.
A network is given 3D joint positions and is asked to predict the rotations from a canonical ``T-pose'' to the input pose.
Predicted rotations are transformed back to joint positions via forward kinematics, and the training loss is on the reconstructed joint positions. We use the code provided by~\cite{hao6d}. Table~\ref{tab:ikstats} presents the results. SVD-Train shows the best overall performance while 6D is closer than in other experiments.
\begin{table}
\caption{\small\textbf{Human pose inverse kinematics}. Following~\protect\cite{hao6d}, we show errors in predicted joint locations  in cm. As above, we show the test errors after training (left), mean errors while training progresses (middle), and percentiles on the right.}
\label{tab:ikstats}
\centering
\resizebox{0.3\textwidth}{!}{
\begin{tabular}{l@{\hskip 0.05in}c@{\hskip 0.04in}c@{\hskip 0.07in}c}
& Mean (cm) & Med & Std \\\cline{2-4}
Euler & 2.59 & 2.04 & 2.08 \\
Axis-Angle & 3.78 & 3.13 & 3.08 \\
Quaternion & 3.09 & 2.44 & 2.55 \\
5D & 2.13 & 1.67 & 1.46 \\
6D & 1.70 & \textbf{1.28} & 1.30 \\
SVD-Train & \textbf{1.61} & 1.29 & \textbf{1.20}
\end{tabular}
}
\begin{tabular}{ll}
\hspace{-1em}
\raisebox{-0.2in}{\includegraphics[width=0.34\textwidth,height=1.3in]{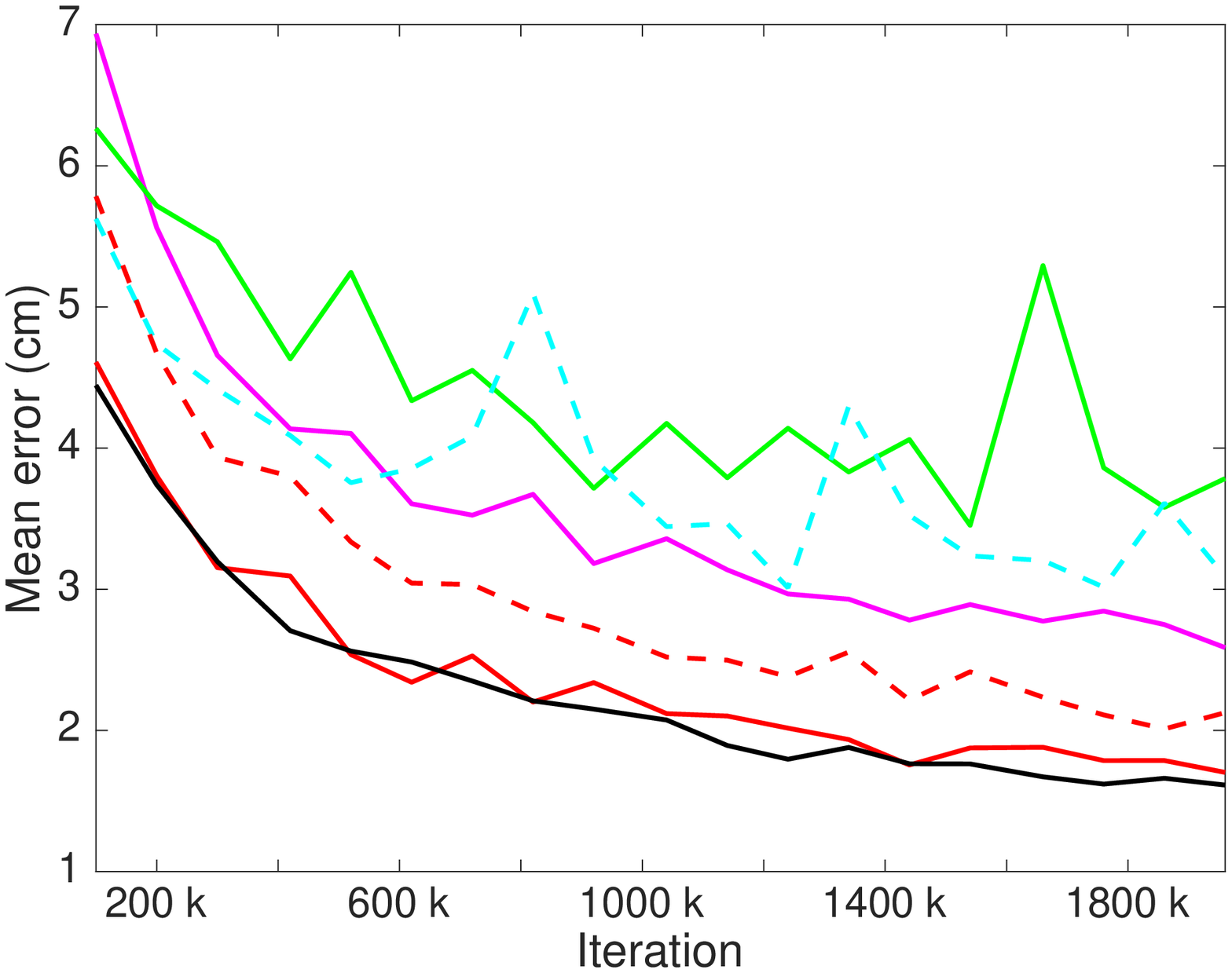}} &
\hspace{-1.5em}
\raisebox{-0.2in}{\includegraphics[width=0.34\textwidth,height=1.3in]{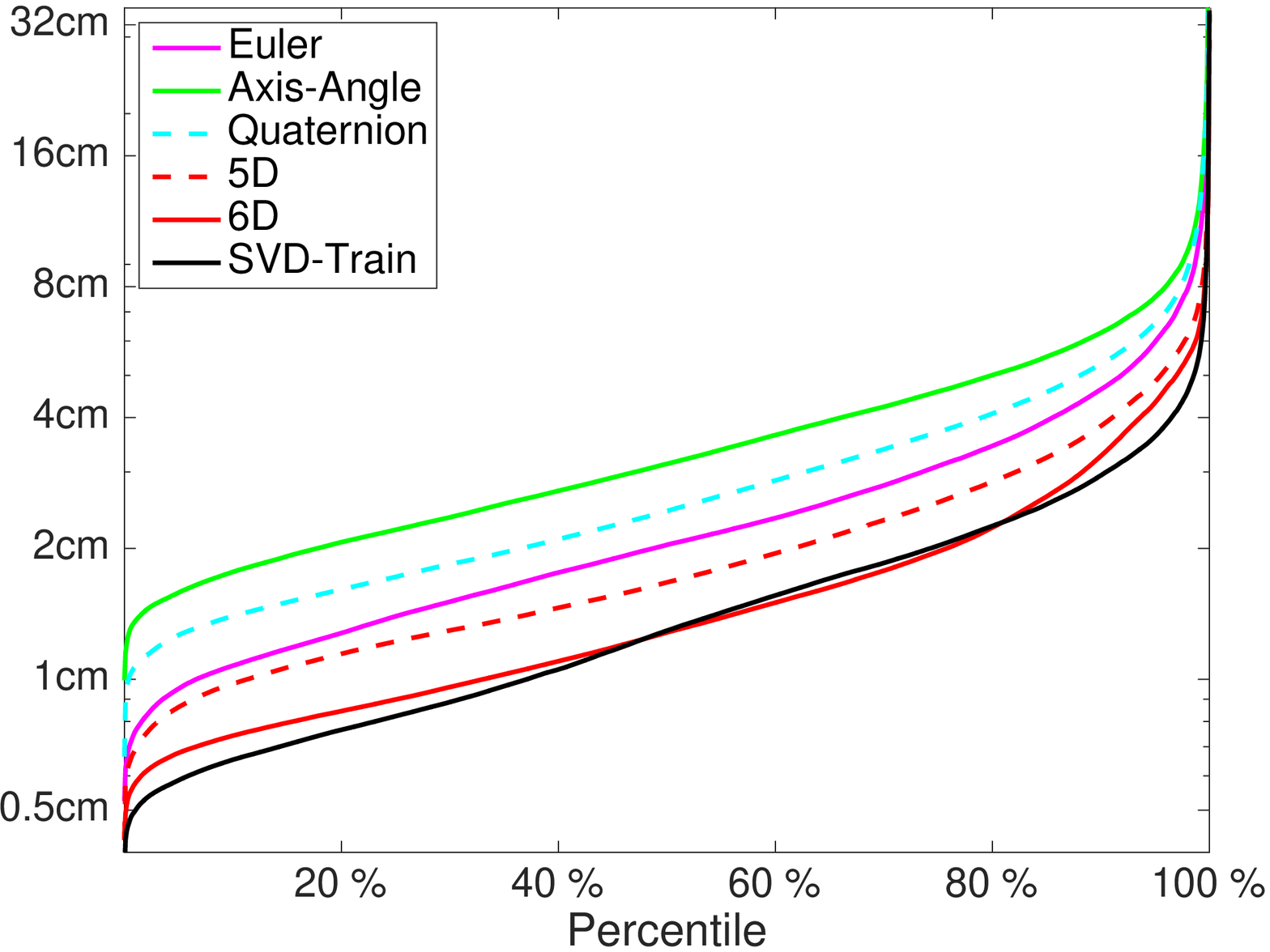}}
\end{tabular}
\end{table}

\subsubsection{Unsupervised depth estimation}\label{sec:depth}
The final experiment considers self-supervised learning of depth and ego-motion from videos~\cite{zhou2017unsupervised}. Given a target image and source images, the model predicts a depth map for the target and camera poses from the target to sources. Source images are warped to the target view using predicted poses, and reconstruction loss on the warped image supervises training. In~\cite{zhou2017unsupervised} the rotational component is parameterized by Euler angles. Following~\cite{zhou2017unsupervised}, we report the single-view depth estimation results on KITTI~\cite{kitti} after 200K steps (Table~\ref{tab:depthestimation}). The error metrics are in meters while accuracy metrics are percentages up to a distance threshold in meters (see~\cite{eigendepth} for a description).

Observe that the difference between the best and second best method in each metric is small. This is not surprising since the camera pose is a small (albeit important) part of a complex deep architecture. Nonetheless, SVD-Train performs best for 4 out of the 7 metrics, and second best in another two. For driving data the motion is likely to be mostly planar for which axis-angle is well suited. Finally, it is worth noting that carefully selecting the rotation representation is important even in more complex models -- the default selection of Euler angles in~\cite{zhou2017unsupervised} is outperformed in every metric.
\begin{table}
\caption{\textbf{Single view depth estimation on KITTI.} We report the same metrics as in~\protect\cite{zhou2017unsupervised}. See Sec.~\ref{sec:depth}.}
\label{tab:depthestimation}
\resizebox{\textwidth}{!}{
\centering
\small
\begin{tabular}{lccccccc}
 & \multicolumn{4}{c}{Error metric $\downarrow$} & \multicolumn{3}{c}{Accuracy metric $\uparrow$}\\
 \cmidrule(lr){2-5}\cmidrule(lr){6-8}
 & Abs Rel	& Sq Rel &	RMSE & RMSE log	& $\delta < 1.25$ & $\delta < 1.25^2$	& $\delta < 1.25^3$	\\
 \cmidrule(lr){2-8}
Euler & 0.216 & 3.163 & 7.169 & 0.291 & 0.720 & 0.893 & 0.952 \\
Axis-Angle & \textbf{0.208} & 2.752 & 7.099 & 0.287 & \textbf{0.723} & 0.894 & \textbf{0.954} \\
Quaternion & 0.218 & 3.055 & 7.251 & 0.294 & 0.707 & 0.888 & 0.950 \\
5D & 0.234 & 4.366 & 7.471 & 0.303 & 0.717 & 0.890 & 0.950 \\
6D & 0.217 & 3.103 & 7.320 & 0.297 & 0.716 & 0.891 & 0.951 \\
SVD-Train & 0.209 & \textbf{2.517} & \textbf{7.045} & \textbf{0.286} & 0.715 & \textbf{0.895} & 0.953
\end{tabular}
}
\end{table}

\section{Conclusion}
The results of the previous sections are broad and conclusive: a continuous 9D unconstrained representation followed by an SVD projection onto $SO(3)$ is consistently an effective, and often the state-of-the-art, representation for 3D rotations in neural networks. It is usable in a variety of application settings including without supervision. The strong empirical evidence is supported by a theoretical analysis that supports SVD as the preferred projection onto $SO(3)$.

\bibliographystyle{plain}
\bibliography{egbib}

\appendix
\section{Complete proof of Proposition 1}\label{sec:propcorrfullproof}
In the main paper, we gave the derivative of the orthogonalization operators $\texttt{SVDO}(M)$ and $\texttt{GS}(M)$ and the resulting error under Gaussian noise, near the identity matrix $M = I$. We now give the complete proof and discussion of Proposition 1 and the additional facts about smoothness of $\texttt{SVDO}(M), \texttt{SVDO}^+(M)$.

Note that since $\texttt{SVDO}(RM) = R\cdot \texttt{SVDO}(M)$ and $\texttt{GS}(RM) = R\cdot \texttt{GS}(M)$ for any orthogonal matrix $R$, and likewise for $\texttt{SVDO}^+, \texttt{GS}^+$ if $R$ is \emph{special} orthogonal. Therefore the error analyses are the same for all matrices $M$:
\begin{equation}
\lVert \texttt{GS}(R + \sigma N) - R \rVert^2_F = \lVert R(\texttt{GS}(I + \sigma R^{-1}N) - I) \rVert^2_F = \lVert \texttt{GS}(I + \sigma N) - I \rVert^2_F
\end{equation}
since orthogonal matrices preserve Frobenius norm and $R^{-1}N$ has the same distribution as $N$ since $N$ was assumed isotropic. (The same applies for the other three functions.)

\emph{Proof of Proposition 1}.
(1) Let $M$ have SVD $M = U \Sigma V^T$ for some orthogonal matrices $U, V$ and diagonal matrix $\Sigma \geq 0$. To first order in $\sigma$, we can expand each of $U, \Sigma, V^T$ as
\begin{align}
U &= U_0(I + \sigma U_1), \\
\Sigma &= \Sigma_0 + \sigma \Sigma_1, \\
V &= V_0(I + \sigma V_1),
\end{align}
with $U_0, V_0$ orthogonal, $U_1, V_1$ antisymmetric and $\Sigma_0, \Sigma_1 \geq 0$ diagonal. This is using the fact that the antisymmetric matrices give the tangent space to the orthogonal matrices. Similarly, the tangent space to the diagonal matrices is given again by the diagonal matrices. This gives an overall expression for $M$ as
\begin{equation} \label{svd-factorization}
M = I + \sigma N = U_0(I + \sigma U_1)(\Sigma_0 + \sigma \Sigma_1)(I + \sigma V_1)^T V_0^T.
\end{equation}
Setting $\sigma = 0$ we see $I = U_0 \Sigma_0 V_0^T,$ which implies $\Sigma_0 = I$ and $U_0 = V_0$. Next, collecting the first-order $\sigma$ terms gives
\begin{equation}
N = U_0(U_1 + \Sigma_1 + V_1^T)U_0^T.
\end{equation}
If a matrix $X$ is (anti-)symmetric and $Q$ is orthogonal, then $QXQ^T$ is again (anti-)symmetric. So, the symmetric and antisymmetric parts of the equation are
\begin{align}
S = U_0 \Sigma_1 U_0^T, \quad A = U_0(U_1 + V_1^T)U_0^T.
\end{align}
Note that the first equation is an SVD of the symmetric part of $N$, while the second equation shows that $U_1$ and $V_1$ satisfy $U_1 + V_1^T = U_0^TAU_0$. Finally, dropping the $\Sigma_0 + \sigma \Sigma_1$ factor from Eq. \eqref{svd-factorization} and expanding out shows that $\texttt{SVDO}(I+\sigma N) = I + \sigma A + O(\sigma^2)$. \smallskip

(2) Let $M = QR$, where $Q$ is orthogonal and $R$ is upper-triangular with positive diagonal. As above, by expanding to first order in $\sigma$ we have
\begin{equation}
I+\sigma N = Q_0(I+\sigma Q_1)(I+\sigma R_1)R_0,
\end{equation}
with $Q_0$ orthogonal, $Q_1$ antisymmetric, and $R_1, R_0$ upper triangular. Setting $\sigma = 0$, we see $I = Q_0 R_0$ and so $Q_0 = R_0 = I$. For the $\sigma$ terms, we split $N$ into its upper, lower and diagonal parts to get
\begin{equation}
U+D+L = Q_1 + R_1,
\end{equation}
which by comparing parts gives $Q_1 = L - L^T$ and $R_1 = U + D + L^T$. Then $\texttt{GS}(M) = I + \sigma(L - L^T)$ by simple algebra.

We now prove Corollary 1.

\paragraph{Corollary 1 (restated).}
If $N$ is $3 \times 3$ with i.i.d. Gaussian entries $n_{ij} \sim \mathcal{N}(0, 1)$, then with error of order $O(\sigma^3)$,

\begin{alignat}{2}
\mathbb{E}[\lVert \emph{\texttt{SVDO}}(M) - I \rVert_F^2] &= 3\sigma^2,\qquad
\mathbb{E}[\lVert \emph{\texttt{GS}}(M) - I \rVert_F^2] &&= 6\sigma^2 \\
\mathbb{E}[\lVert \emph{\texttt{SVDO}}(M) - M \rVert_F^2] &= 6\sigma^2,\qquad
\mathbb{E}[\lVert \emph{\texttt{GS}}(M) - M\rVert_F^2] &&= 9\sigma^2
\end{alignat}

\emph{Proof}. Simplifying the error expressions using the first-order calculations in the Proposition gives
\begin{eqnarray}
\lVert \texttt{SVDO}(M) - I \rVert_F^2 &=& \lVert \sigma A \rVert^2_F, \\
\lVert \texttt{GS}(M) - I \rVert_F^2 &=& \lVert \sigma (L - L^T) \rVert^2_F, \\
\lVert \texttt{SVDO}(M) - M \rVert_F^2 &=& \lVert -\sigma S \rVert^2_F, \\
\lVert \texttt{GS}(M) - M\rVert_F^2 &=& \lVert -\sigma(U+D+L^T)  \rVert^2_F,
\end{eqnarray}
with notation for $S, A, U, D, L$ as in the proposition. Thus each expression is $\sigma^2$ times the Frobenius norm of the corresponding matrix. Each entry of $A, L-L^T, S$ and $U+D+L^T$ is a linear combination of the entries of $N$, hence is Gaussian since $N$ has i.i.d.\ Gaussian entries $n_{ij} \sim \mathcal{N}(0, 1)$. The expectations are the sums of the entrywise expectations of these matrices. For example, $A = \tfrac{1}{2}(N - N^T)$ has six nonzero entries of the form $\tfrac{1}{2}(n_{ij} - n_{ji})$, each having variance $\tfrac{1}{2}$, so $\mathbb{E}[\lVert A \rVert^2_F] = 3$. For $L - L^T$, the above diagonal entries are $-n_{ji}$ and the below-diagonal entries are $n_{ij}$, and the diagonal is $0$, so the total variance is $6$. The other two calculations are similar (the entries do not all have the same variances).

{\bf Remark.}
The tangent space to the identity matrix along the orthogonal matrices is the space of antisymmetric matrices. Both of the calculations above can be thought of as giving orthogonal approximations of the form
\begin{equation}
I + \sigma N \approx I + \sigma A',
\end{equation}
where $A'$ is a choice of antisymmetric matrix that depends on the approximation method. The fact that $\texttt{SVDO}(M)$ produces the approximation $A' = A = \tfrac{1}{2}(N - N^T)$ means it corresponds to the natural projection of $N$ onto the orthogonal tangent space. By contrast, $\texttt{GS}(M)$ produces $A' = L - L^T$, essentially a "greedy" choice with respect to the starting matrix (minimizing the change to the leftmost columns). For certain matrices $\texttt{GS}$ can have smaller error: for example if $N$ happens to be upper-triangular, $\texttt{GS}(M) = I$ and the error is zero. For isotropic noise, however, the SVD approximation is the most efficient in expectation.

\subsection[Accuracy of error estimates for large noise]{Accuracy of error estimates as $\sigma$ increases}
\begin{figure}[h]
\centering
\includegraphics[width=0.4\textwidth]{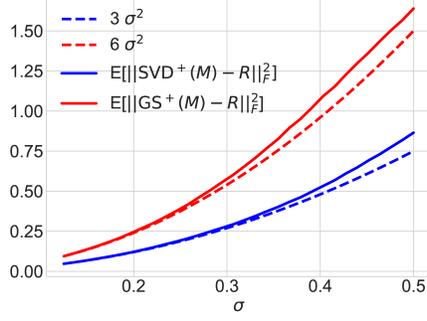}
\caption{\small\textbf{Simulations.} We plot our derived approximations against numerical simulations of the expected error in reconstruction under additive noise. For each $\sigma$ we compute the numerical expectation with 100K trials. These plots can provide a sanity check of our derivations.}
\label{fig:simulations}
\end{figure}

From Corollary 1 (Sec 3.3) we see special-orthogonalization with Gram-Schmidt ($\texttt{GS}^+$) produces twice the error in expectation as SVD ($\texttt{SVDO}^+$) for $SO(3)$ reconstruction when inputs are perturbed by Gaussian noise. We compare these derived errors with numerical simulations. See Figure~\ref{fig:simulations}.  

\section{Proof of smoothness and discussion}\label{sec:appsmoothness}
\begin{prop}
The symmetric orthogonalization $\texttt{SVDO}(M)$ is a smooth function of $M$ if $\det(M) \ne 0$. \end{prop}

\emph{Proof.}
We use the Implicit Function Theorem and the least-squares characterization of $\texttt{SVDO}(M)$ as
\begin{equation}
    \texttt{SVDO}(M) = \argmin_{Q \in O(n)} \lVert M - Q\rVert_F^2. \label{o-least-squares}
\end{equation}
We calculate the derivative with respect to $Q \in O(n)$: for $A$ an antisymmetric matrix, 
\begin{equation}
\lim_{\epsilon \to 0} \tfrac{1}{\epsilon}(\lVert M - Q(I+\epsilon A)\rVert_F^2 - \lVert M - Q\rVert_F^2) = -2 \Tr(M^TQA).
\end{equation}
If this vanishes for every $A$, then $M^TQ$ is symmetric, that is, $(M, Q)$ is a root of the function $g(M, Q) = M^TQ - Q^TM$. Let $M_0$ be a fixed matrix. As discussed above, the optimal solution to this equation is given by an SVD, $M_0 = U_0 S_0 V_0^T$, yielding $Q_0 = U_0V_0^T$. To show that $Q$ is a smooth function of $M$, it suffices by the Implicit Function Theorem to show that the Jacobian matrix $\frac{\partial g}{\partial Q}$ is full-rank at $(M_0, Q_0)$. To see this, we differentiate it again:
\begin{equation}
\frac{\partial g}{\partial Q}(A) = \lim_{\epsilon \to 0} \frac{1}{\epsilon}(g(M_0, Q_0(I+\epsilon A)) - g(M_0, Q_0)) = M_0^TQ_0A - A^T Q_0^T M_0,
\end{equation}
where $A$ is antisymmetric. Some algebra shows that this is, equivalently,
\begin{equation}
\frac{\partial g}{\partial Q}(A) = V_0(S_0 V_0^TAV_0 + V_0^T A V_0 S_0)V_0^T.
\end{equation}
To see that this is an invertible transformation of $A$, note that conjugating by $V_0$ is invertible since $V_0$ is orthogonal. So it is equivalent to show that the function
\begin{equation}
A \mapsto S_0 A + A S_0    
\end{equation}
is invertible. This function just rescales the entry $a_{ij}$ to $(s_i + s_j)a_{ij}$. Since the singular values are positive this is invertible as desired.

\begin{prop}
The special symmetric orthogonalization is a smooth function of $M$ if either of the following is true:
\begin{itemize}
    \item $\det(M) > 0$,
    \item $\det(M) < 0$ and the smallest singular value of $M$ has multiplicity one.
\end{itemize}
\end{prop}

\emph{Proof sketch}. The analysis is identical to the main proof, except that if $\det(M) < 0$, $S_0$ is effectively altered so that the last entry is changed from $s_n$ to $-s_n$. Thus the function $A \mapsto S_0 A + A S_0$ now sends the entry $a_{ij}$ to $(\pm s_i \pm s_j) a_{ij}$, with negative signs at $i=n$ and/or $j = n$. If $s_n$ occurred with multiplicity one, the result is still invertible since $s_i - s_n \ne 0$ for $i \ne n$ and for $i= j = n$ the coefficient is $-2s_n$. Otherwise, however, $s_{n-1} = s_n$ and the operation sends $a_{n-1, n}$ to $(s_{n-1} - s_n)a_{n-1,n} = 0$; likewise for $a_{n, n-1}$. In this case there are many optimal \emph{special} orthogonalizations of $M_0$, and the operation is not even continuous in a neighborhood of $M_0$.

\section{Gradients}\label{sec:appgrads}
Here we provide the detailed derivation sketched out in Sec~\ref{backprop}. 
We will first analyze \pgrad{M} for $\texttt{SVDO}(M)$. With $\circ$ denoting the Hadamard product, from~\cite{ionesculong,townsendsvd}
we have
\begin{eqnarray}\small
\pgrad{M} & = & U [(F \circ (U^T \pgrad{U} - \pgrad{U}^T U))\Sigma + \Sigma ( F \circ (V^T \pgrad{V} - \pgrad{V}^T V))]V^T, \label{gradstartapp}\\
F_{i,j} & = & 
\begin{cases}
    \frac{1}{s_i^2 - s_j^2},& \text{if } i \neq j \label{invsigapp}\\
    0,& \text{if } i = j
\end{cases}, \;\;s_i = \Sigma_{ii}.
\end{eqnarray}
Letting $X = U^T \pgrad{U} - \pgrad{U}^T U$, and $Y = V^T \pgrad{V} - \pgrad{V}^T V$,  we see that $X, Y$ are antisymmetric. Furthermore, since $\lVert \texttt{SVDO}(M)\ - R\rVert_F^2 = 2\Tr(\mathbb{I}_n) - 2\Tr(UV^TR^T)$, then $\pgrad{U} = -2RV$, and $\pgrad{V} = -2R^TU$. This leads directly to $X=Y^T=-Y$.
We can simplify Eq.~\ref{gradstartapp} as
\begin{eqnarray}
\pgrad{M} & = & U((F \circ X) \Sigma - \Sigma (F \circ X)) V^T.
\end{eqnarray}
Inspecting the individual elements of $(F \circ X) \Sigma$ and $\Sigma (F \circ X))$ we have
\begin{eqnarray}
\left((F \circ X) \Sigma\right)_{ij} = \frac{X_{ij} s_j}{s_i^2 - s_j^2}, \qquad
\left(\Sigma (F \circ X)\right)_{ij} = \frac{X_{ij} s_i}{s_i^2 - s_j^2}. 
\end{eqnarray}
Letting $Z = (F \circ X) \Sigma - \Sigma (F \circ X)$, we can simplify $\pgrad{M} = UZV^T$ where the elements of $Z$ are
\begin{eqnarray}\small
Z_{ij} & = &
\begin{cases}
    \frac{-X_{ij}}{s_i + s_j} ,& \text{if } i \neq j \label{svdpgrad}\\
    0,& \text{if } i = j.
\end{cases}
\end{eqnarray}
For $\texttt{SVDO}(M)$ Eq.~\ref{svdpgrad} tells us $\pgrad{M}$ is undefined whenever two singular values are both zero and large when their sum is very near zero. 

For $\texttt{SVDO}^+(M)$, if $\det(M) > 0$ then the analysis is the same as above. If $\det(M) < 0$, the extra factor $D = \mathrm{diag}(1, 1, \ldots, -1)$ effectively changes the smallest singular value $s_n$ to $-s_n$. The derivation is otherwise unchanged. In particular the denominator in equation \eqref{svdpgrad} is now $s_j - s_n$ or $s_n - s_i$ if either $i$ or $j$ is $n$.%

\subsection{Gradients observed during training}
\begin{figure}[h]
\centering
\resizebox{1.0\textwidth}{!}{
\begin{tabular}{l@{}r}
\hspace*{-0.25in}
\includegraphics[width=0.6\textwidth]{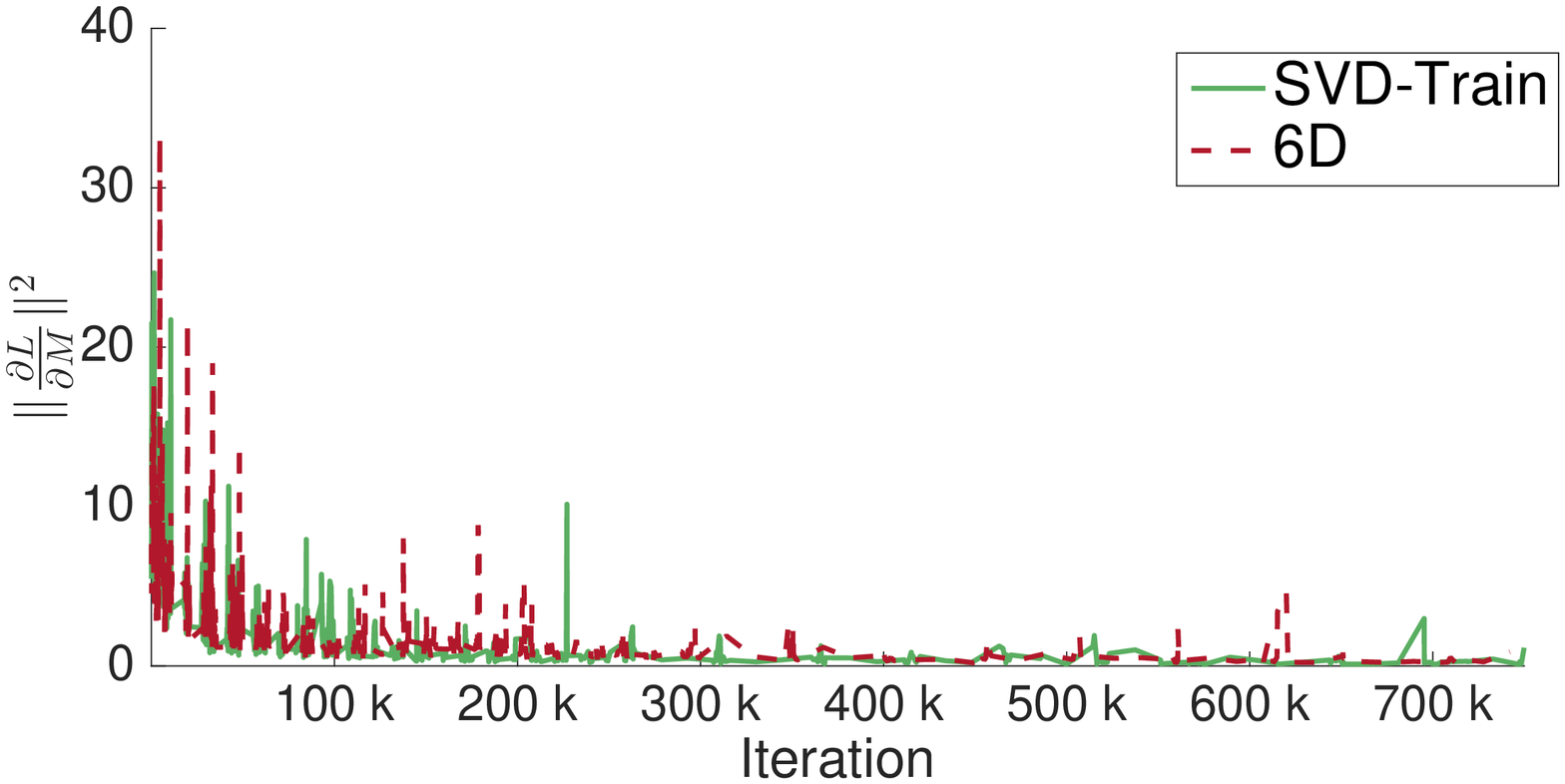} & \includegraphics[width=0.6\textwidth]{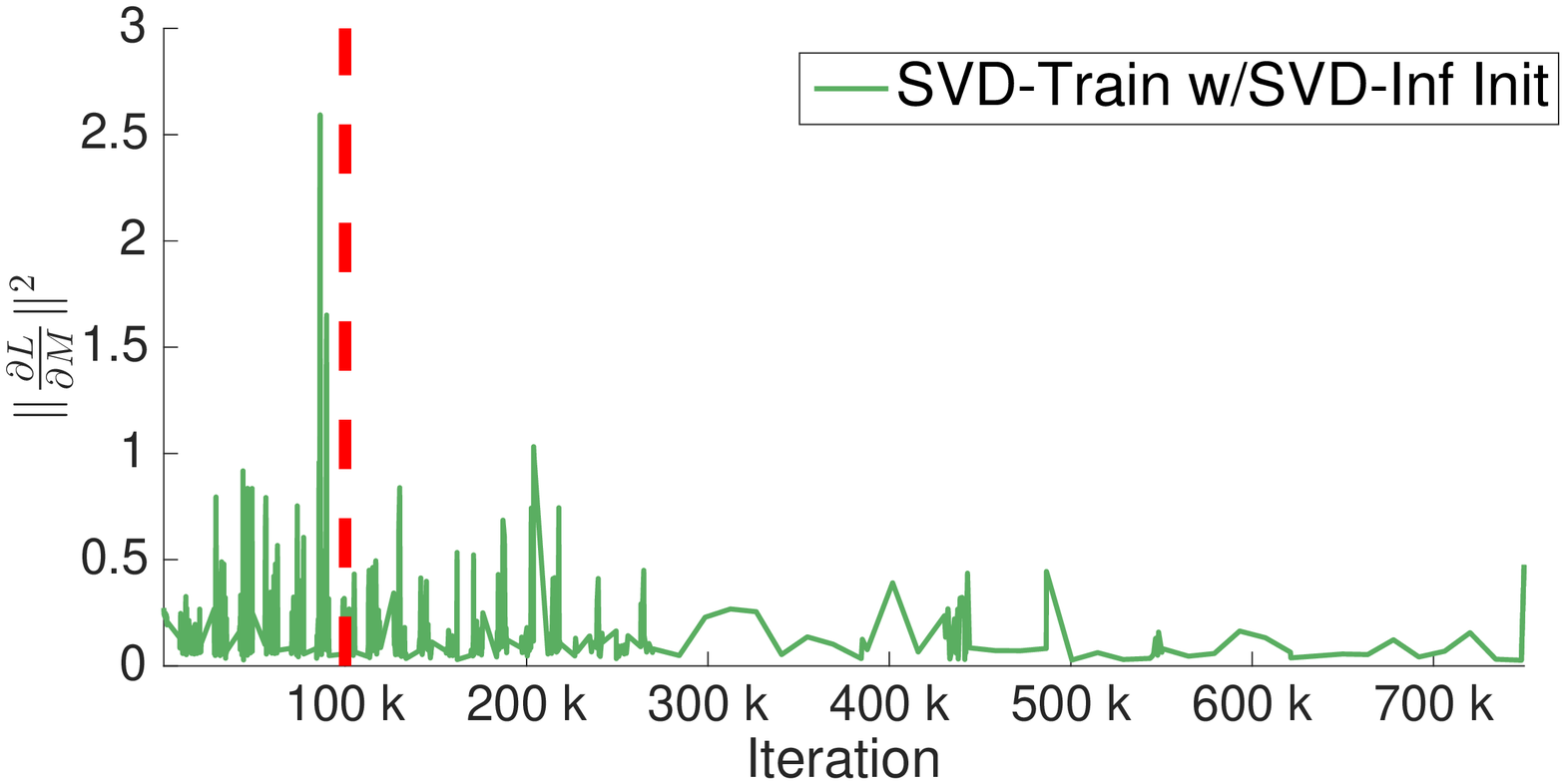}
\end{tabular}
}
\caption{\small\textbf{Gradients.} Left are the gradient norms $\lVert \frac{\partial L}{\partial M} \rVert_F^2$ for the point cloud alignment experiment. SVD-Train and 6D have similar profiles. On the right the network is trained with SVD-Inf for the first 100K steps, then SVD-Train. During the first 100K steps the network learns to output matrices close to $SO(3)$ and this eliminates sources of high gradient norms in Eq.~\ref{svdpgrad}.}
\label{fig:grads}
\end{figure}

In Figure~\ref{fig:grads} (left) we see the gradient norms observed while training for point cloud alignment (Section~\ref{sec:expptcloud1}). SVD-Train has the same profile as for 6D ($\texttt{GS}^+$). SVD-Train converges quickly (relative to all other methods) in all of our experiments, indicating no instabilities due to large gradients.

On the right of Figure~\ref{fig:grads} we profile the gradients for the scenario where we begin training with the SVD-Inference loss and switch to SVD-Train after 100K steps (after roughly 4\% of training iterations). SVD-Inf trains the network to produce outputs that are close to $SO(3)$, which eliminates some conditions of instability in Eq.~\ref{svdpgrad}. This is confirmed by seeing much smaller gradient norms after switching to SVD-Train at 100K steps. Note, this approach was never used (or needed) in our experiments.

\section{Experiments}\label{sec:appexp}
\subsection{Additional baseline implementation details}\label{sec:appbaselines}
\begin{itemize}
    \item \textbf{6D} is the partial Gram-Schmidt method which computes $\texttt{GS}^+(M)$. Our implementation follows exactly the code provided by~\cite{hao6d}.
    \item \textbf{5D} maps outputs in $\mathbb{R}^5$ onto $SO(3)$ as described in ~\cite{hao6d}.
    \item \textbf{Spherical Regression} (\textbf{$S^2$-Reg}) regresses to $n$-spheres. Following~\cite{s2reg}, we use regression to $S^1$ for Pascal3D+ and $S^3$ regression (quaternions) for their ModelNet experiments (section 4.2). The method combines abs.\ value regression with sign classification. Our implementation of the final regression layers follows the provided code. We select the hyperparameter that balances the classification and regression losses by a simple line search in the neighborhood of the default provided in ~\cite{s2reg}.
    
    $S^2$-Reg uses both classification and regression losses, not surprisingly we were unable to train successfully on any of the unsupervised rotation experiments. The closest we came was on unsupervised point cloud alignment (Sec 4.4). With careful hyperparameter tuning the model completed training with mean test errors near $90^\circ$.
    
    \item \textbf{3D-RCNN}~\cite{rcnn3d} combines likelihood estimation and regression (via expectation) for predicting Euler angles. This representation also requires both classification and regression losses for training, and we were unable to make the model train successfully on the unsupervised rotation experiments.
    
    \item \textbf{Geodesic-Bin-and-Delta} (\textbf{M$_G$}~\cite{classificationregression2018}) combines classification over quantized pose space (axis-angle representation) with regression of offsets. For our experiments with where observed rotations are uniformly distributed over $SO(3)$ (Sec.~\ref{sec:expptcloud1}, ~\ref{sec:mnet}), $K$-means clustering is ineffective.
    Instead we quantize $SO(3)$ by uniformly sampling a large number (1000) of rotations (larger values did not improve results). We found this version of Geodesic-Bin-and-Delta outperformed the One-delta-network-per-pose-bin variation in these experiments. For Pascal3D+ we follow the reference and use $K$-means with $K=200$. This method also requires both classification and regression losses and we were unable to train successfully in the unsupervised setting. 
    
    \item \textbf{Quaternion}, \textbf{Euler angles}, and \textbf{axis-angle} are the classic parameterizations. In each case they are converted to matrix form before the loss is applied. In our experiments we did not filter any outputs from the network representing angles (e.g.\ clipping values or applying activations such as $\operatorname{tanh}$). We found this gave the best results overall.
\end{itemize}

\subsection{Learning rate decay}
An observation from the point cloud registration results is that the curves for mean test errors as training progresses do not decay smoothly as one might expect for any method (Table 1, middle, in the main paper). This is in part due to the training code from ~\cite{hao6d} does not utilize a learning rate decay for this experiment. It is reasonable to observe the variance in evaluation when a decay is introduced as would be common in practice.  Table~\ref{tab:geodecay} (left) shows the curves when the learning rate is exponentially decayed (decay rate of 0.95, decay steps of 35K). The evaluation over time is smoother but the results are consistent with those presented in the main paper.

\subsection{Geodesic loss}
In ~\cite{hao6d} it was shown that geodesic loss for training did not alter the results much, and we have the same observation. Table~\ref{tab:geodecay} (right) shows the geodesic loss results.
\begin{table}[h]
\caption{\small\textbf{Left: Training point cloud alignment with learning rate decay}. Evaluation is smoother over time but the comparative analysis does not change. \textbf{Right: geodesic loss.} Training with geodesic loss for point cloud alignment. Relative performances are consistent with squared Frobenius loss (Table~\ref{tab:PTstats} in main paper).}
\label{tab:geodecay}
\centering
\hspace*{-0.15in}
\begin{tabular}{l}
\includegraphics[width=0.55\textwidth]{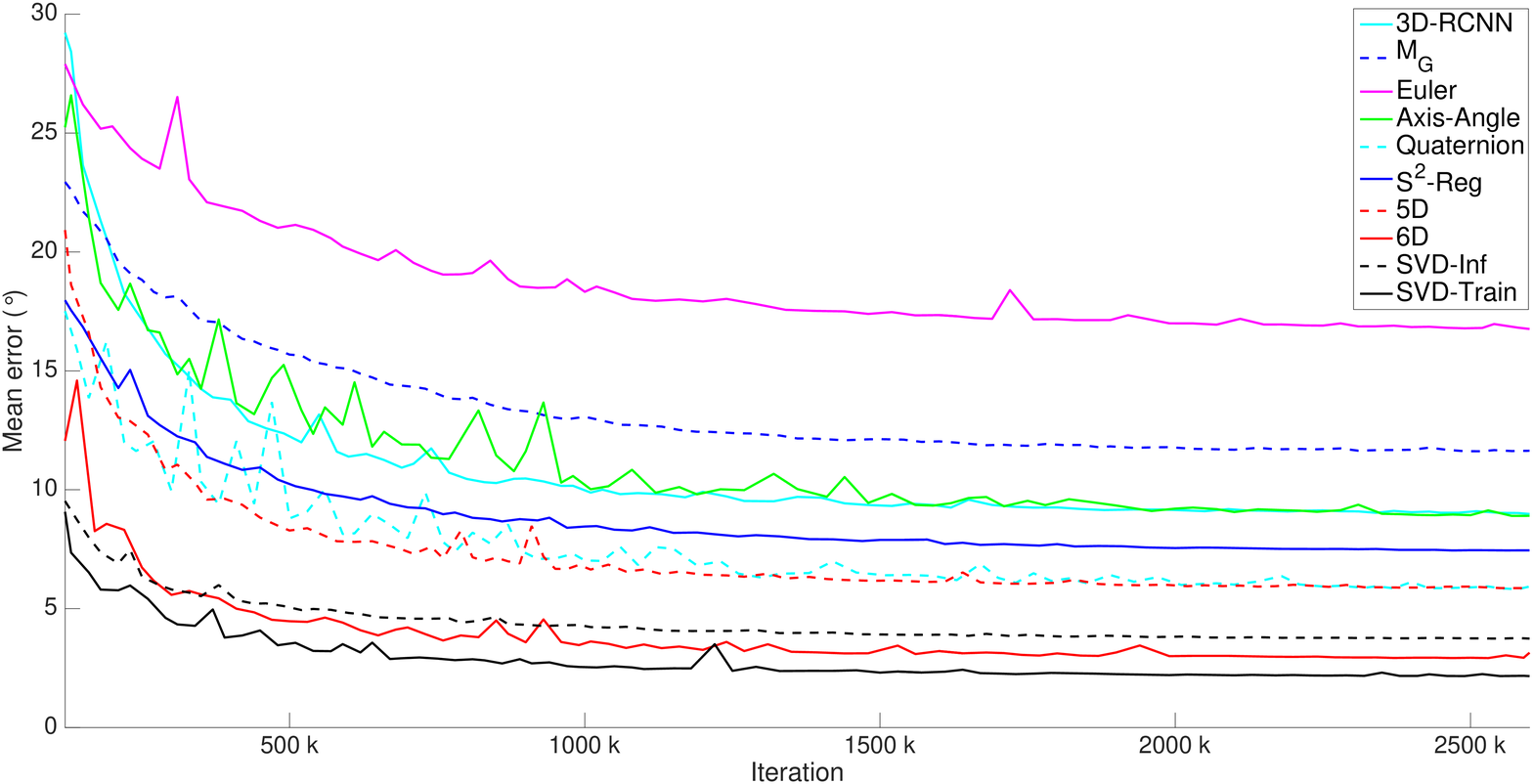}
\end{tabular}
\resizebox{0.4\textwidth}{!}{
\begin{tabular}{l@{\hskip 0.05in}c@{\hskip 0.04in}c@{\hskip 0.07in}c}
& Mean ($^\circ$) & Med & Std \\\cline{2-4}
5D & 3.88 & 2.08 &  9.19 \\
6D & 2.29 & 1.30 &  7.52 \\
SVD-Train & 2.05 & 1.28 &  7.15
\end{tabular}
}
\end{table}

\subsection{Architecture details}
\subsubsection{3D point cloud alignment (Sec.~\ref{sec:expptcloud1})}\label{ptarch}
The model architecture follows exactly the architecture described in~\cite{hao6d}. Point clouds are embedded with simplified PointNet (4-layer MLP) ending with a global max-pooling. Three dense layers make up the regression network. The output dimensionality of the final layer depends on the chosen representation. For classification+regression models the final layers follows the relevant references.
\subsubsection{3D Pose estimation from 2D images (Sec.~\ref{sec:mnet}), and Pascal3D+ (Sec.~\ref{sec:p3d})}
The regression layers are the same as above. The image embeddings are generated with MobileNet~\cite{howard2017mobilenets}. We found no difference in performance between MobileNet and VGG16~\cite{vgg} (not surprising given the comparison in~\cite{howard2017mobilenets}).
\subsection{Inverse kinematics (Sec.~\ref{sec:ik}) and unsupervised depth estimation (Sec.~\ref{sec:depth})}
For these experiments the training and evaluation code is provided by ~\cite{hao6d} and ~\cite{zhou2017unsupervised}, respectively. We simply change the rotation representation layer.

\subsection{Remark on timings}
We noticed no measurable difference in training time with $\texttt{SVDO}^+$ as expected, since the forward and backward pass for small, e.g.\ $3\times 3$, SVD is efficient.

\subsection{Pascal3D+ full results.}\label{sec:apppascal}
Here in Table~\ref{tab:pascal3dall} we show the results for all 12 categories in the Pascal3D+.
\begin{table}[ht]
\caption{\textbf{Pascal 3D+.} Results for all 12 categories.}
\label{tab:pascal3dall}
\resizebox{\textwidth}{!}{
\centering
\small
\begin{tabular}{lcccccccccccccccc}
& \multicolumn{4}{c}{aeroplane} & \multicolumn{4}{c}{bottle} & \multicolumn{4}{c}{chair} & \multicolumn{4}{c}{sofa} \\
 \cmidrule(lr){2-5}\cmidrule(lr){6-9}\cmidrule(lr){10-13}\cmidrule(lr){14-17}
 & \multicolumn{3}{c}{Accuracy@} &  Med$^\circ$ & \multicolumn{3}{c}{Accuracy@} &  Med$^\circ$ & \multicolumn{3}{c}{Accuracy@} & Med$^\circ$ & \multicolumn{3}{c}{Accuracy@} &  Med$^\circ$\\
 & 10$^\circ$ & 15$^\circ$ & 20$^\circ$ & Err & 10$^\circ$ & 15$^\circ$ & 20$^\circ$ & Err & 10$^\circ$ & 15$^\circ$ & 20$^\circ$ & Err & 10$^\circ$ & 15$^\circ$ & 20$^\circ$ & Err \\
\cmidrule(lr){2-4}\cmidrule(lr){5-5}\cmidrule(lr){6-8}\cmidrule(lr){9-9}\cmidrule(lr){10-12}\cmidrule(lr){13-13}\cmidrule(lr){14-16}\cmidrule(lr){17-17}
3D-RCNN & 32.8 & 52.5 & 77.9 & 13.5 & 61.3 & 74.2 & 90.3 & 7.2 & 29.7 & 45.1 & 69.8 & 17.2 & 37.1 & 54.3 & 80.0 & 14.2 \\
$M_G$ & 22.1 & 45.1 & 82.4 & 16.0 & 48.4 & 62.9 & 87.1 & 11.0 & 23.1 & 45.6 & 75.8 & 15.9 & 31.4 & 51.4 & 74.3 & 14.4 \\
Euler & 15.2 & 35.3 & 70.1 & 19.8 & 58.1 & 69.4 & 91.9 & 9.0 & 9.3 & 28.6 & 58.8 & 25.3 & 22.9 & 45.7 & 77.1 & 16.3 \\
Axis-Angle & 16.7 & 34.8 & 74.5 & 20.0 & 50.0 & 67.7 & 91.9 & 10.4 & 11.5 & 27.5 & 69.8 & 21.7 & 11.4 & 40.0 & 80.0 & 16.3 \\
Quaternion & 28.9 & 46.6 & 77.5 & 16.0 & 53.2 & 71.0 & 91.9 & 8.3 & 19.8 & 37.4 & 73.1 & 18.6 & 34.3 & 62.9 & 77.1 & 11.7 \\
$S^2$-Reg & 46.6 & 67.6 & 87.3 & 10.6 & 56.5 & 69.4 & 91.9 & 8.8 & 37.4 & 61.5 & 84.6 & 12.7 & 37.1 & 65.7 & 85.7 & 11.2 \\
5D & 21.6 & 38.7 & 75.5 & 17.3 & 54.8 & 66.1 & 93.5 & 9.2 & 17.6 & 34.6 & 72.0 & 19.1 & 17.1 & 54.3 & 77.1 & 14.2 \\
6D & 24.0 & 42.6 & 75.5 & 17.3 & 54.8 & 71.0 & 95.2 & 9.3 & 20.9 & 39.6 & 78.6 & 17.2 & 34.3 & 54.3 & 88.6 & 13.3 \\
SVD-Inf & 26.0 & 57.4 & 86.3 & 13.3 & 56.5 & 75.8 & 95.2 & 8.9 & 20.3 & 43.4 & 77.5 & 16.8 & 45.7 & 60.0 & 88.6 & 11.0 \\
SVD-Train & 22.1 & 43.6 & 77.0 & 17.4 & 53.2 & 75.8 & 93.5 & 7.7 & 24.2 & 39.0 & 71.4 & 17.6 & 40.0 & 57.1 & 85.7 & 12.7 \\
\newline & & & & & & & & & & & & & & & & \\
& \multicolumn{4}{c}{bicycle} & \multicolumn{4}{c}{bus} & \multicolumn{4}{c}{diningtable} & \multicolumn{4}{c}{train} \\
 \cmidrule(lr){2-5}\cmidrule(lr){6-9}\cmidrule(lr){10-13}\cmidrule(lr){14-17}
 & \multicolumn{3}{c}{Accuracy@} &  Med$^\circ$ & \multicolumn{3}{c}{Accuracy@} &  Med$^\circ$ & \multicolumn{3}{c}{Accuracy@} & Med$^\circ$ & \multicolumn{3}{c}{Accuracy@} &  Med$^\circ$\\
 & 10$^\circ$ & 15$^\circ$ & 20$^\circ$ & Err & 10$^\circ$ & 15$^\circ$ & 20$^\circ$ & Err & 10$^\circ$ & 15$^\circ$ & 20$^\circ$ & Err & 10$^\circ$ & 15$^\circ$ & 20$^\circ$ & Err \\
\cmidrule(lr){2-4}\cmidrule(lr){5-5}\cmidrule(lr){6-8}\cmidrule(lr){9-9}\cmidrule(lr){10-12}\cmidrule(lr){13-13}\cmidrule(lr){14-16}\cmidrule(lr){17-17}
3D-RCNN & 17.8 & 38.6 & 72.3 & 16.9 & 88.7 & 91.5 & 93.7 & 4.4 & 46.7 & 60.0 & 66.7 & 12.2 & 65.7 & 74.7 & 82.8 & 6.4 \\
$M_G$ & 11.9 & 31.7 & 66.3 & 20.9 & 76.1 & 88.0 & 95.1 & 7.6 & 26.7 & 53.3 & 60.0 & 12.8 & 48.5 & 66.7 & 82.8 & 10.1 \\
Euler & 9.9 & 20.8 & 68.3 & 23.4 & 47.2 & 66.9 & 87.3 & 10.5 & 26.7 & 40.0 & 73.3 & 16.6 & 42.4 & 63.6 & 80.8 & 11.1 \\
Axis-Angle & 13.9 & 31.7 & 70.3 & 21.3 & 38.7 & 69.7 & 93.7 & 12.0 & 26.7 & 53.3 & 80.0 & 14.4 & 40.4 & 64.6 & 82.8 & 11.6 \\
Quaternion & 15.8 & 30.7 & 67.3 & 22.4 & 69.7 & 83.8 & 92.3 & 7.4 & 33.3 & 46.7 & 73.3 & 17.3 & 56.6 & 68.7 & 81.8 & 8.7 \\
$S^2$-Reg & 21.8 & 45.5 & 75.2 & 16.1 & 93.7 & 98.6 & 99.3 & 3.8 & 33.3 & 46.7 & 66.7 & 15.3 & 66.7 & 76.8 & 84.8 & 6.2 \\
5D & 10.9 & 26.7 & 68.3 & 21.1 & 52.1 & 72.5 & 93.0 & 9.6 & 33.3 & 60.0 & 66.7 & 11.4 & 35.4 & 49.5 & 78.8 & 15.5 \\
6D & 14.9 & 27.7 & 71.3 & 22.0 & 66.9 & 83.8 & 94.4 & 7.9 & 13.3 & 46.7 & 73.3 & 15.3 & 63.6 & 73.7 & 80.8 & 7.7 \\
SVD-Inf & 10.9 & 33.7 & 84.2 & 19.0 & 80.3 & 92.3 & 95.8 & 6.1 & 53.3 & 60.0 & 73.3 & 10.0 & 58.6 & 73.7 & 82.8 & 8.5 \\
SVD-Train & 9.9 & 26.7 & 80.2 & 20.9 & 67.6 & 85.2 & 96.5 & 7.9 & 33.3 & 53.3 & 73.3 & 13.0 & 63.6 & 72.7 & 81.8 & 8.4 \\
\newline & & & & & & & & & & & & & & & & \\
& \multicolumn{4}{c}{boat} & \multicolumn{4}{c}{car} & \multicolumn{4}{c}{motorbike} & \multicolumn{4}{c}{tvmonitor} \\
 \cmidrule(lr){2-5}\cmidrule(lr){6-9}\cmidrule(lr){10-13}\cmidrule(lr){14-17}
 & \multicolumn{3}{c}{Accuracy@} &  Med$^\circ$ & \multicolumn{3}{c}{Accuracy@} &  Med$^\circ$ & \multicolumn{3}{c}{Accuracy@} & Med$^\circ$ & \multicolumn{3}{c}{Accuracy@} &  Med$^\circ$\\
 & 10$^\circ$ & 15$^\circ$ & 20$^\circ$ & Err & 10$^\circ$ & 15$^\circ$ & 20$^\circ$ & Err & 10$^\circ$ & 15$^\circ$ & 20$^\circ$ & Err & 10$^\circ$ & 15$^\circ$ & 20$^\circ$ & Err \\
\cmidrule(lr){2-4}\cmidrule(lr){5-5}\cmidrule(lr){6-8}\cmidrule(lr){9-9}\cmidrule(lr){10-12}\cmidrule(lr){13-13}\cmidrule(lr){14-16}\cmidrule(lr){17-17}
3D-RCNN & 12.6 & 23.2 & 52.6 & 27.0 & 65.5 & 76.8 & 86.3 & 6.7 & 24.6 & 46.5 & 81.6 & 15.6 & 35.5 & 53.9 & 82.9 & 13.2 \\
$M_G$ & 16.8 & 27.4 & 56.8 & 25.2 & 51.2 & 70.2 & 86.9 & 9.8 & 15.8 & 40.4 & 79.8 & 17.1 & 23.0 & 46.1 & 77.0 & 15.9 \\
Euler & 1.1 & 5.3 & 28.4 & 46.4 & 25.0 & 50.6 & 79.8 & 14.5 & 13.2 & 30.7 & 67.5 & 21.3 & 23.0 & 46.7 & 79.6 & 15.7 \\
Axis-Angle & 4.2 & 13.7 & 42.1 & 35.0 & 21.4 & 53.6 & 82.1 & 14.0 & 15.8 & 32.5 & 73.7 & 19.6 & 25.7 & 42.8 & 81.6 & 16.6 \\
Quaternion & 9.5 & 23.2 & 54.7 & 27.1 & 45.2 & 64.9 & 86.9 & 10.5 & 18.4 & 32.5 & 78.9 & 18.9 & 25.7 & 51.3 & 81.6 & 14.3 \\
$S^2$-Reg & 18.9 & 42.1 & 66.3 & 16.7 & 70.2 & 85.7 & 98.2 & 7.7 & 28.9 & 56.1 & 86.8 & 13.6 & 38.8 & 56.6 & 78.9 & 13.3 \\
5D & 4.2 & 10.5 & 48.4 & 32.1 & 23.2 & 46.4 & 84.5 & 16.1 & 9.6 & 30.7 & 79.8 & 20.3 & 22.4 & 36.8 & 69.1 & 18.8 \\
6D & 12.6 & 18.9 & 52.6 & 29.0 & 44.0 & 67.3 & 89.3 & 11.4 & 11.4 & 36.8 & 88.6 & 17.2 & 30.9 & 50.7 & 85.5 & 14.7 \\
SVD-Inf & 17.9 & 31.6 & 56.8 & 23.3 & 56.5 & 76.2 & 91.1 & 8.9 & 21.9 & 47.4 & 86.8 & 15.6 & 30.3 & 52.6 & 86.2 & 14.3 \\
SVD-Train & 13.7 & 25.3 & 52.6 & 25.4 & 42.3 & 63.1 & 85.7 & 11.4 & 18.4 & 40.4 & 81.6 & 18.3 & 32.9 & 50.0 & 86.2 & 14.6
\end{tabular}
}
\end{table}

\end{document}